\documentclass[twoside,11pt]{article}

\usepackage[preprint]{jmlr2e}

\usepackage[T1]{fontenc}    \usepackage{url}            \usepackage{booktabs}       \usepackage{amsfonts}       \usepackage{nicefrac}       \usepackage{microtype}      

\usepackage{amsmath}
\usepackage{mathtools}
\usepackage{thm-restate}
\usepackage{algorithm}
\usepackage{algorithmic}
\usepackage{wrapfig}

\usepackage{setspace}
\usepackage{xspace}

\usepackage[usenames,dvipsnames]{xcolor}
\usepackage{wrapfig}
\usepackage{array} \usepackage{multirow}
\usepackage{enumitem}
\usepackage{mdframed}
\usepackage{relsize}
\usepackage{caption}
\usepackage{verbatim}

\usepackage[nameinlink]{cleveref}

\newcommand{\eps}{\varepsilon}
\renewcommand{\epsilon}{\varepsilon}

\renewcommand{\tilde}{\widetilde}
\renewcommand{\bar}{\overline}

\def\:#1{\protect \ifmmode {\mathbf{#1}} \else {\textbf{#1}} \fi}

\newcommand{\X}{\mathcal{X}}
\newcommand{\Y}{\mathcal{Y}}

\newcommand{\bzero}{\mathbf{0}}

\newcommand{\bI}{\mathbf{I}}

\newcounter{cnt-lem-quad-variation}
\newcommand{\calS}{\mathcal{S}}

\renewcommand{\epsilon}{\varepsilon}

\newcommand{\norm}[1]{\left\Vert #1 \right\Vert}

\newcommand{\E}{{\mathbb E}}

\newcommand{\Ex}{\mathcal{E}}
\newcommand{\hh}{\mathcal{H}}

\newcommand{\scal}[2]{\left\langle{#1},{#2}\right\rangle}

\newcommand{\R}{\mathbb{R}}
\newcommand{\N}{\mathbb{N}}

\newcommand{\nt}[1]{{\bf \color{red} #1}}

\newcommand{\Hcal}{\mathcal{Z}}
\newcommand{\Lcal}{\mathcal{L}}
\newcommand{\Scal}{\mathcal{S}}
 \newcommand{\rev}[1]{{#1}}

\usepackage{lastpage}

\firstpageno{1}

\begin{document}

\title{On Mixup Regularization}

\author{\name Luigi Carratino\thanks{This work was done in part at Google Brain, Paris.} \email luigi.carratino@dibris.unige.it \\
       \addr MaLGa - University of Genova, Italy
       \AND
       \name Moustapha Ciss\'e \email moustaphacisse@google.com \\
       \addr Google Research - Brain team, Accra
       \AND
       \name Rodolphe Jenatton \email rjenatton@google.com \\
       \addr Google Research - Brain team, Berlin
       \AND
       \name Jean-Philippe Vert\thanks{Now at Owkin, Paris, France} \email jean-philippe.vert@m4x.org \\
       \addr Google Research - Brain team, Paris
       }

\maketitle

\begin{abstract}Mixup is a data augmentation technique that creates new examples as convex combinations of training points and labels. 
  This simple technique has empirically shown to improve the accuracy of many state-of-the-art models in different settings and 
  applications, but the reasons behind this empirical success remain poorly understood. 
  In this paper we take a substantial step in explaining the theoretical foundations of Mixup, 
  by clarifying its regularization effects. 
  We show that Mixup can be interpreted as standard empirical risk minimization estimator subject to 
  a combination of data transformation and random perturbation of the transformed data. 
  We gain two core insights from this new interpretation.
  First, the data transformation suggests that, at test time, a model trained with Mixup should also be applied to transformed data, a one-line change in code that we show empirically to improve both accuracy and calibration of the prediction.
  Second, we show how the random perturbation of the new interpretation of Mixup
  induces multiple known regularization schemes, 
  including label smoothing and reduction of the Lipschitz constant of the estimator. 
  These schemes interact synergistically with each other, 
  resulting in a self calibrated and effective regularization effect that prevents overfitting 
  and overconfident predictions. 
  We corroborate our theoretical analysis with experiments that support our conclusions.
\end{abstract}

\section{Introduction}

Regularization is an essential component of machine learning models and 
plays an even more important role in deep learning \citep{goodfellow2016deep}. 
Regularization mechanisms can take various forms. They can be \emph{explicitly} enforced by: 
(i) applying various penalties to the parameters of the models~\citep{hinton1987learning, krogh1992simple, bartlett2017spectrally,neyshabur2015path, sedghi2018singular,arjovsky2017wasserstein}, (ii) injecting noise to the internal representations of the network~\citep{srivastava2014dropout, gal2016dropout} and/or to its outputs~\citep{szegedy2016rethinking,Muller2019When},
or (iii) normalizing the activations~\citep{he2016deep,salimans2016weight}.
Or they can be \emph{implicit} thanks to: (j) parameter sharing in architectures such as convolutional 
networks~\citep{lecun1998gradient}, (jj) the choice of the optimization algorithm~\citep{neyshabur2017implicit}, e.g., stochastic gradient descent converging to small norm solutions~\citep{arora2019implicit}, 
or (jjj) through data augmentation and transformation \citep{goodfellow2016deep}.
There is a large body of work explaining the effects of the numerous explicit and implicit 
regularization procedures existing in the literature. 
For instance, explicit regularization schemes usually proceed from analysis aiming to control specific 
characteristics of a model such as robustness~\citep{hein2017formal,cisse2017parseval} or 
calibration~\citep{guo2017calibration,Muller2019When}, while the forms of implicit regularization are 
often understood through the angle of generalization~\citep{neyshabur2017implicit, arora2019implicit}. 
However, the regularization effects of modern data augmentation procedures are less theoretically understood. 

\begin{figure}
    \centering
    \includegraphics[width=0.8\textwidth]{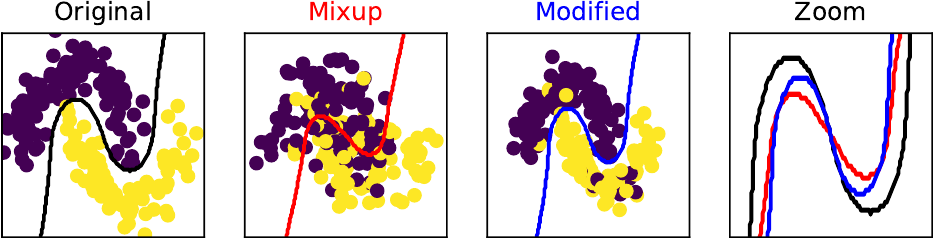}
    \caption{\small
        Illustration of how training a model with Mixup (second plot) differs from training 
        a model on original data (first plot), the fourth plot highlighting the discrepancy between 
        the Bayes classifiers in both situations (black vs red). 
        To explain this difference, we show in this paper that the model trained with Mixup 
        can be interpreted as a regularized version of a model trained on modified data 
        (third plot, blue curve on the zoom plot), and characterize both the data modification 
        (from black to blue)  and the regularization effect (from blue to red). 
        Both effects interact synergistically to confer Mixup strong regularization properties, 
        which may explain its good empirical behavior in a variety of tasks.
        }
    \label{fig:overview}
    \vspace{-1\baselineskip}
\end{figure}

Data augmentation is a core ingredient for successful deep learning pipelines. 
It helps to alleviate sample size issues and prevent overfitting. 
In simple cases, there are known equivalences between data augmentation and other existing explicit 
regularization procedures, e.g., training with additional noisy points in least-squares regression 
is equivalent to Tikhonov regularization~\citep{bishop1995training}. 
Similar analysis have recently been performed to explain the regularization effect of 
dropout~\citep{srivastava2014dropout,wager2013dropout,wei2020implicit}.
In this work, we focus on \textit{Mixup}~\rev{\citep{Zhang2018mixup,tokozume2018between}}, a recently introduced data-augmentation 
technique that consists in generating examples as random convex combinations of data points and labels 
from the training set (as illustrated in Figure~\ref{fig:overview}). 
Despite its simplicity, Mixup has been shown to substantially improve generalization on a broad 
range of tasks ranging from computer vision~\rev{\citep{Zhang2018mixup,tokozume2018between}} to natural language 
processing~\citep{guo2020nonlinear} and semi-supervised learning~\citep{berthelot2019mixmatch}. 
The success of Mixup has triggered several variations such as adaptive Mixup~\citep{guo2019mixup}, 
manifold Mixup~\citep{verma2018manifold} and Cutmix~\citep{yun2019cutmix}, but the reasons why Mixup and 
its variants work so well in practice remain poorly understood.

Mixup's primary motivation was to alleviate overfitting in training deep neural networks~\citep{Zhang2018mixup}. 
However, previous studies have also empirically noticed other desirable regularization effects it induces. 
These include improved calibration~\citep{thulasidasan2019mixup}, robustness to input adversarial noise~\citep{Zhang2018mixup}, 
and robustness to label corruption~\citep{Zhang2018mixup}. \citet{Zhang2018mixup} also 
showed it helps stabilize notoriously difficult learning problems such as generative adversarial networks. 
Traditionally, separate regularization methods are applied to induce the above effects. 
For example, label smoothing~\citep{szegedy2016rethinking,Muller2019When} leads to better calibration, 
while dropout improves generalization~\citep{srivastava2014dropout,wager2013dropout} and robustness to 
label corruption~\citep{arpit2017closer}.  Lipschitz regularization helps stabilize the training of generative 
adversarial networks~\citep{arjovsky2017wasserstein,gulrajani2017improved}. It also leads to increased robustness to 
adversarial perturbations~\citep{hein2017formal,cisse2017parseval}. 
\Cref{tab:comparisons} shows a comparison of various regularization procedure proposed in the literature, 
and the effect they are known to induce on the model. 
Although all these desirable regularization effects have been observed empirically, 
no theoretical explanation has been given yet.

In this work, we propose the first theoretical analysis of Mixup\footnote{\rev{After we published a first version of this work~\citep{Carratino2020Mixup}, \citet{Zhang2021HowDM} independently derived a similar and complementary analysis of Mixup; we summarize in \Cref{sec:discussion} the main differences between both works.}} to better understand the reasons for its empirical success. 
We show that Mixup can be analyzed through the lenses of empirical risk minimization with random perturbations, 
and exploit ingredients from previous analysis of dropout~\citep{wager2013dropout, khalfaoui2019asni, wei2020implicit} 
to derive a regularized objective function that sharply captures the regularization effects of Mixup. 
In particular, our analysis sheds some light on the multiple effects that Mixup borrows from the popular 
regularization mechanisms listed above such as label smoothing~\citep{pereyra2017regularizing} (output noise) or  
dropout~\citep{srivastava2014dropout} (input noise), and how it uniquely combines them to improve calibration and 
smooth the Jacobian of the model.
We further show that this analysis points out a missing step in learning with Mixup, and we present 
how applying a simple transformation when evaluating at test time the function learned with Mixup
can improves accuracy and calibration.
More precisely, we make the following contributions (illustrated in Figure~\ref{fig:overview}):
\begin{itemize}\item We show that Mixup can be reinterpreted as a standard empirical risk minimization procedure, applied to a transformation of the original data perturbed by random perturbations, and give explicit formulas for the data transformation and the perturbations.
\item In particular, we show that the Mixup transformations shrinks both the inputs and the outputs towards their mean, the later creating a form of regularization by label smoothing. We notably give a formal description of the effect of label smoothing in the case of the cross-entropy loss where it translates into an increase in the entropy of the predictions.
\item We show that Mixup learns functions from a modified version of the input space of the training points to a modified version of the output space of the training points.
Thus, we present how to properly evaluate the learned functions to further improve accuracy and calibration. 
\item We characterize the random perturbations induced by Mixup on both the inputs and the outputs, as well as their dependency and their correlation structure.
\item We deduce an approximation of the regularization induced by Mixup, and highlight in particular how it regularizes both the model and its derivatives. We discuss in details the specific cases of classification with cross-entropy loss, and least squares regression.
\item We provide empirical support for our interpretation of Mixup regularization.
\end{itemize}

\begin{table}[!htbp]
\begin{center}
\begin{scriptsize}
\setlength\tabcolsep{4pt}
\begin{tabular}{@{}l|c|c|c|c|c|@{}}
\toprule
Method & Calibration  & Jacobian Reg.  & Robustness Label Noise  & Input Normaliz.  \\
\midrule
Label smooth.~\citep{szegedy2016rethinking}       &     $\checkmark$        &   &     &    \\
Spectral Reg.~\citep{cisse2017parseval}   &           &  $\checkmark$           &      &\\
Dropout~\citep{wager2013dropout}       & $\checkmark$  &  $\checkmark$  & $\checkmark$     &      \\
Temperat. scaling~\citep{guo2017calibration} & $\checkmark$   &   &    & \\
Mixup~\citep{Zhang2018mixup} & $\checkmark$   & $\checkmark$  &   $\checkmark$   & $\checkmark$\\
\bottomrule
\end{tabular}
\end{scriptsize}
\end{center}
\caption{Summary of the effects induced by various regularizers. Absence of checkmark means the corresponding effect is not known for this regularizer. \nt{}}
\label{tab:comparisons}
\end{table}

The rest of the paper is organized as follows. In \Cref{sec:notations}, we introduce notations used throughout the paper and describe the setting of empirical risk minimization and learning with Mixup. In \Cref{sec:erm}, we show how Mixup can be interpreted as an empiricial risk minimization on modified data with random perturbations. In \Cref{sec:regu} we analyze the regularization effect of Mixup through a quadratic Taylor approximation of the formulation derived in \Cref{sec:erm}. In \Cref{sec:discussion}, we discuss in detail several aspects of Mixup that the theoretical analysis in \Cref{sec:regu} and \Cref{sec:erm} suggest, and confront them to experimental validations. The proofs of all results are detailed in the Annex, together with additional experimental results.

 \section{Notations and setting}\label{sec:notations}

\paragraph{Notations.}
For any $n\in\N$, $[n] = \{1,\dots,n\}$ is the set of nonzero integers up to $n$,
$\:1_n \in\R^n$ is the $n$-dimensional vector of ones, 
and $\bzero_n$ and $\bI_n\in\R^{n\times n}$ are the $n$-dimensional null and identity matrices, respectively.
For any two matrices $Z,Z'$ of equal size we note $\scal{Z}{Z'} = \text{Trace}(Z^\top Z')$ their 
Frobenius inner product, and with $\|Z\|_{{F}} = \sqrt{\scal{Z}{Z}}$ the Frobenius norm.
For any vector $x\in\R^n$, matrix $M \in \R^{n\times m}$ and positive semi-definite matrix 
$Z\in\R^{n \times n}$, we denote by $\norm{x}^2_Z = x^\top Z x$ 
the squared semi-norm of $x$ with metric $Z$, and with $\norm{M}^2_Z = \scal{M}{ZM} = \text{Trace}(M^\top Z M)$
the squared Frobenius norm with metric $Z$. 
For any function $f: \R^a \rightarrow \R^b$ and vector $x\in \R^a$,
we denote respectively by $\nabla f(x) \in \R^{b\times a}$ and $\nabla^2 f(x) \in \R^{b\times a \times a}$ the Jacobian and Hessian of $f$ at $x$, i.e., if $f(x) = (f_1(x_1,\ldots,x_a), \ldots, f_b(x_1,\ldots,x_a))$, then $[\nabla f(x)]_{i,j} = \partial f_i / \partial x_j (x)$ and $[\nabla^2 f(x)]_{i,j,k} = \partial^2 f_i / \partial x_j \partial x_k (x)$, for $(i,j,k)\in[b]\times[a]\times[a]$. Note in particular that if $f:\R^a\rightarrow \R$, then the gradient of $f$ is a row vector $\nabla f(x)\in\R^{1\times a}$. When $f$ has several arguments and we wish to take partial derivatives with respect to some of the arguments, we explicitly name the different arguments as $f(u,v)$ and then indicate as a subscript to the $\nabla$ sign the argument(s) according to which we take derivatives, e.g., if $u\in\R^{a_u}$ and $v\in\R^{a_v}$, then $\nabla_{u}f(u,v)\in\R^{b\times a_u}$ is the Jacobian of $f$ with respect to $u$, and $\nabla^2_{uv} f \in \R^{b\times a_u \times a_v}$ is the tensor of second derivatives of $f$ of the form $[\nabla^2_{uv} f(u,v)]_{i,j,k} = \partial^2 f_i / \partial u_j \partial v_k (u,v)$ for $(i,j,k)\in[b]\times[a_u]\times[a_v]$. We recall that if $f:\R^{a_u + a_v}\rightarrow\R$ is twice continuously differentiable, then $\nabla_{uv}f = \nabla_{vu}f^\top$, by Schwarz's theorem. For any random variable $X$ and measurable function $f$, we denote by $\E_X f(X)$ the expectation of $f(X)$, or simply $\E f(X)$ when no confusion is possible. For any shape parameters $\alpha, \beta>0$, and any interval $[a,b]\subset [0,1]$, $\text{Beta}_{[a,b]}(\alpha,\beta)$ denotes the truncated Beta distribution on $[a,b]$, i.e., the distribution of a random variable with values in $[a,b]$ and density proportional to $x^{\alpha-1} (1-x)^{\beta-1}$ on $[a,b]$. We simply write $\text{Beta}(\alpha,\beta)=\text{Beta}_{[0,1]}(\alpha,\beta)$ for the usual Beta distribution. For any $p\in[0,1]$, $\text{Ber}(p)$ denotes the Bernoulli distribution with parameter $p$. For any $c\in\N$, we denote by 
\begin{equation*}
\Delta_c = \{u \in \R^c; u^\top \mathbf{1}_c = 1\ \ \text{and for}\ j\in[c], \ u_j \geq 0\}
\end{equation*}
the simplex in $\R^c$, and for any $p \in \Delta_c$, we denote by
$\Hcal(p) = -\sum_{j=1}^c p_j \log(p_j)$
the entropy of a categorical distribution with parameter $p$.

\paragraph{Learning problem.}
We consider a training set $S_n = \{(x_1,y_1), \dots, (x_n,y_n)\}$ made of $n$ input/output pairs, where for each pair $i\in[n]$, $x_i \in \X \subset \R^d$ and $y_i \in \Y \subset \R^c$. This covers in particular the regression or binary classification settings, where $c=1$, or the multivariate regression and multiclass classification setting, where $y_i$ is an embedding of the class of $x_i$ in $\R^c$, e.g., the one-hot encoding by taking $c$ equal to the total number of classes and letting $y_i\in \{0,1\}^c$ be the binary vector with all entries equal to zero except for the one corresponding to the class of $x_i$. We further denote the mean input and output as
$$
\bar{x} = \frac{1}{n}\sum_{i=1}^n x_i\,,\quad \quad \bar{y} = \frac{1}{n}\sum_{i=1}^n y_i\,,
$$
and the empirical variance and covariance matrices 
or inputs and outputs as
$$
\Sigma_{xx} = \frac{1}{n}\sum_{i=1}^n (x_i - \bar{x})(x_i - \bar{x})^\top\,,\quad \Sigma_{xy} = \frac{1}{n}\sum_{i=1}^n (x_i - \bar{x})(y_i - \bar{y})^\top\,,\quad \Sigma_{yy} = \frac{1}{n}\sum_{i=1}^n (y_i - \bar{y})(y_i - \bar{y})^\top\,.
$$
Our goal is to learn from $S_n$ a function \rev{$f:\X \rightarrow \R^c$} to predict the output corresponding to any new input $x\in\X$ via \rev{$\rho(f(x))$, where $\rho:\R^c \rightarrow \Y$ maps an $\R^c$-valued prediction to an element of $\Y$; standard mappings include the identity $\rho(y)=y$ for regression problems, and the softmax operator $\rho(y)_i = e^{y_i} / \left(\sum_{j=1}^c e^{y_j}\right)$ for multiclass classification problems}. For that purpose, we formulate the inference problem as an optimization problem:
\begin{equation}\label{eq:minrisk}
\min_{f\in \hh} \Ex(f)\,,
\end{equation}
where $\hh$ is a class of candidate functions, such as linear functions or deep neural networks, and $\Ex(f)$ is a risk functional that depends on $S_n$. The most standard risk used in machine learning is the empirical risk, defined for any loss function \rev{$\ell:\Y\times\R^c\rightarrow \R$} by:
\begin{equation}\label{eq:erm}
\Ex^{\text{Empirical}}(f)= \frac{1}{n}\sum_{i=1}^n \ell(y_i, f(x_i))\,.
\end{equation}
Solving (\ref{eq:minrisk}) with the empirical risk (\ref{eq:erm}) is often called \emph{empirical risk minimization} (ERM), and is typically performed in practice by first-order numerical optimization such as stochastic gradient descent~\citep{Bottou2008Tradeoffs}. Standard losses $\ell$ include the squared error (in regression) and the cross-entropy loss \rev{applied to the softmax mapping} (in classification, assuming that $\forall y\in\Y, y^\top \:1_c=1$, which is true for one-hot encoded classes and their convex combinations):
\begin{equation}\label{eq:loss}
\forall(y,u)\in\rev{\Y\times\R^c}\,,\quad \ell^{\text{SE}}(y,u) = \frac{1}{2}\|y-u\|^2\,,\quad\quad \ell^{\text{CE}}(y,u) =  \log\left( \sum_{i=1}^c e^{u_i}\right) - y^\top u\,.
\end{equation}

\paragraph{Mixup.}
Instead of minimizing the empirical risk (\ref{eq:erm}),  Mixup \citep{Zhang2018mixup}  creates new random input/output samples by taking convex combinations of pairs of training samples, and minimizes the corresponding empirical risk. With our notations, Mixup therefore minimizes the following \emph{Mixup risk} over $f\in\hh$:
\begin{equation}\label{eq:mixup}
\Ex^{\text{Mixup}}(f)= \frac{1}{n^2}\sum_{i=1}^n \sum_{j=1}^n \E_\lambda \ell\left(\lambda y_i + (1-\lambda)y_j, f(\lambda x_i + (1-\lambda) x_j) \right)\,,
\end{equation}
where $\lambda \sim \text{Beta}(\alpha,\alpha)$, and $\alpha$ is a parameter of Mixup. The minimization of (\ref{eq:mixup}) is typically performed by stochastic gradient descent, where $\lambda$ is sampled at each iteration to obtain a stochastic gradient. In practice, \citet{Zhang2018mixup} suggest to sample minibatches of training pairs, and generate Mixup random pairs within the minibatch, which also produces a stochastic gradient of (\ref{eq:mixup}). \section{Mixup as a perturbed ERM}\label{sec:erm}

The Mixup risk (\ref{eq:mixup}) is defined as a sum over pairs of samples, making a comparison with standard ERM approaches (\ref{eq:erm}) not direct. The following result shows that the Mixup risk can be equivalently rewritten as a standard empirical risk, over modified input/output pairs (as in the third plot of Figure~\ref{fig:overview}), subject to random perturbations.

\begin{theorem}\label{thm:mixupAsErm}
Let $\theta \sim \text{Beta}_{[\frac{1}{2},1]}(\alpha,\alpha)$ and $j \sim \text{Unif}([n])$ 
be two random variables with $\alpha >0$, $n> 0$ and let $\bar{\theta} = \E_\theta \theta$. For any training set $\calS_n$, let $(\tilde{x}_i,\tilde{y}_i)$ for any $i\in[n]$ be the modified input/output pair given by
\begin{equation}\label{eq:modif}
\begin{cases}
\tilde{x}_i &= \bar{x} + \bar{\theta}(x_i - \bar{x})\,,\\
\tilde{y}_i &= \bar{y} + \bar{\theta}(y_i - \bar{y})\,,
\end{cases}
\end{equation}
and $(\delta_i,\epsilon_i)$ be the random perturbations given by:
\begin{equation}\label{eq:perturbations}
\begin{cases}
\delta_i &= (\theta - \bar{\theta}) x_i + (1-\theta) x_j - (1-\bar{\theta}) \bar{x}\,, \\
\epsilon_i &= (\theta - \bar{\theta}) y_i + (1-\theta) y_j - (1-\bar{\theta}) \bar{y}\, .
\end{cases}
\end{equation}
Then for any $i\in[n]$, $\E_{\theta,j} \delta_i = \E_{\theta,j} \epsilon_i = 0$, and for any function $f\in\hh$,
\begin{equation}\label{eq:mixupERM}
\Ex^{\text{Mixup}}(f)= \frac{1}{n} \sum_{i=1}^n \E_{\theta,j} \ell\left( \tilde{y}_i + \epsilon_i, f(\tilde{x}_i + \delta_i)\right)\,.
\end{equation}
\end{theorem}
Both $\delta_i$ and $\epsilon_i$ are random vectors because they are functions of $\theta$ and $j$ in (\ref{eq:perturbations}), which are themselves random variables. We hence use the notation $\E_{\theta,j}$ in (\ref{eq:mixupERM}). Note also $\bar{\theta}\in[1/2, 1]$ meaning that the transformation from $(x_i,y_i)$ to $(\tilde{x}_i,\tilde{y}_i)$ in (\ref{eq:modif}) shrinks the inputs and the outputs towards their mean.

\Cref{thm:mixupAsErm}
and the expression (\ref{eq:mixupERM}) of the Mixup risk allow us to re-interpret Mixup as a combination of two standard techniques: (i) transforming each input/output pair $(x_i,y_i)$ into $(\tilde{x}_i,\tilde{y}_i)$, and (ii) adding zero-mean random perturbations $(\delta_i,\epsilon_i)$ to each transformed pair, before minimizing the empirical risk. 
This helps us to understand the effects of training a model with Mixup by studying each technique and their interaction.
In particular, perturbing input data is a classical approach to regularize ERM estimators~\citep{bishop1995training,srivastava2014dropout,wager2013dropout,wei2020implicit}, and we study in detail in the next section the particular regularization induced by the Mixup perturbations on both inputs and outputs, before interpreting the resulting regularization aspects of Mixup due to both data transformation and perturbation in~\Cref{sec:discussion}. \section{The regularization effects of Mixup}\label{sec:regu}

We now study the effect of the random perturbations $(\delta_i,\epsilon_i)$ for $i\in[n]$ in the Mixup risk (\ref{eq:mixupERM}). While perturbing inputs with additive or multiplicative noise (e.g., dropout), and independently perturbing outputs (resulting, e.g., in label smoothing) have been widely studied, the Mixup perturbation (\ref{eq:mixupERM}) is unique in the sense that it is applied to both inputs and outputs simultaneously, and that the input and output perturbations are not independent from each other by (\ref{eq:perturbations}). 
In order to study the regularization effect of these perturbations, we first characterize the covariance structure among the input and output perturbations.
\begin{lemma}\label{lem:corr}
Let $\bar{\theta}$ and $\sigma^2$ be respectively the mean and variance of a $\text{Beta}_{[\frac{1}{2},1]}(\alpha,\alpha)$ distributed random variable, and $\gamma^2 = \sigma^2 + (1-\bar{\theta})^2$. For any $i\in[n]$, let
\begin{equation}\label{eq:sigmai}
    \begin{split}
        \Sigma_{\tilde{x}\tilde{x}}^{(i)} &= \frac{\sigma^2 (\tilde{x}_i - \bar{x}) (\tilde{x}_i-\bar{x})^\top + \gamma^2 \Sigma_{\tilde{x}\tilde{x}}}{\bar{\theta}^2}\,,\\
        \Sigma_{\tilde{y}\tilde{y}}^{(i)} &= \frac{\sigma^2 (\tilde{y}_i - \bar{y}) (\tilde{y}_i-\bar{y})^\top + \gamma^2 \Sigma_{\tilde{y}\tilde{y}}}{\bar{\theta}^2}\,,\\
        \Sigma_{\tilde{x}\tilde{y}}^{(i)} &= \frac{\sigma^2 (\tilde{x}_i - \bar{x}) (\tilde{y}_i-\bar{y})^\top + \gamma^2 \Sigma_{\tilde{x}\tilde{y}}}{\bar{\theta}^2} \,.
    \end{split}
\end{equation}
Then, for any $i\in[n]$, the random perturbations defined in (\ref{eq:perturbations}) satisfy
\begin{equation}\label{eq:cov}
\E_{\theta,j}\delta_i \delta_i^\top = \Sigma_{\tilde{x}\tilde{x}}^{(i)}\,,
\quad
\E_{\theta,j}\epsilon_i \epsilon_i^\top = \Sigma_{\tilde{y}\tilde{y}}^{(i)} \,, 
\quad \text{and}\quad
\E_{\theta,j}\delta_i \epsilon_i^\top = \Sigma_{\tilde{x}\tilde{y}}^{(i)} \,.
\end{equation}
\end{lemma}

Following recent lines of work that interpret various random perturbations such as dropout as regularization \citep{wager2013dropout,wei2020implicit}, we can now introduce and study an approximate Mixup risk:
\begin{equation}\label{eq:mixupQ}
\Ex^{\text{Mixup}}_Q(f)= \frac{1}{n} \sum_{i=1}^n \E_{\theta,j} \ell_Q^{(i)}\left( \tilde{y}_i + \epsilon_i, f(\tilde{x}_i + \delta_i)\right)\,,
\end{equation}
obtained by replacing the loss function $\ell(\tilde{y},f(\tilde{x}))$ by a second-order quadratic Taylor approximation near each modified input/output training pairs $(\tilde{x}_i,\tilde{y}_i)$, namely, for any $i\in[n]$ and $(\delta,\epsilon)\in\X\times\Y$:
\begin{equation}\label{eq:taylor}
\begin{split}
    \ell_Q^{(i)}\left( \tilde{y}_i + \epsilon, f(\tilde{x}_i + \delta)\right)
    &= \ell\left( \tilde{y}_i, f(\tilde{x}_i )\right) + \nabla_y \ell \left( \tilde{y}_i, f(\tilde{x}_i )\right) \epsilon + \nabla_u \ell \left( \tilde{y}_i, f(\tilde{x}_i )\right) \nabla_x f(\tilde{x}_i )\delta\\
    &+ \frac{1}{2} \scal{\delta\delta^\top}{\nabla f(\tilde{x}_i)^\top \nabla^2_{uu}\ell(\tilde{y}_i,f(\tilde{x}_i)) \nabla f(\tilde{x}_i) + \nabla_u\ell(\tilde{y}_i,f(\tilde{x}_i))\nabla^2 f(\tilde{x}_i)}\\
&+ \frac{1}{2} \scal{\epsilon \epsilon^\top}{\nabla^2_{yy}\ell(\tilde{y}_i,f(\tilde{x}_i))}
+ \scal{\epsilon \delta^\top}{\nabla^2_{yu}\ell(\tilde{y}_i,f(\tilde{x}_i))\nabla f(\tilde{x}_i)}\,,
\end{split}
\end{equation}
assuming both $\ell$ and $f$ are twice continuously differentiable. Due to its quadratic form as a function of input and output perturbations, the approximate Mixup risk (\ref{eq:mixupQ}) can be re-expressed as a regularized ERM risk, as shown in the next result. We note that the expression we derive is in fact valid for \emph{any} joint perturbation of the inputs and outputs with covariance structure given in (\ref{eq:cov}).

\begin{theorem}\label{thm:mixupreg}
For any twice continuously differentiable loss $\ell(y,u)$, the approximate Mixup risk at any twice differentiable $f\in\hh$ satisfies
\begin{equation}\label{eq:mixapprox}
 \Ex^{\text{Mixup}}_Q(f)= \frac{1}{n} \sum_{i=1}^n  \ell( \tilde{y}_i , f(\tilde{x}_i)) + R_1(f) + R_2(f) + R_3(f) + R_4(f)\,,
\end{equation}
where
\begin{align*}
    R_1(f) &= \frac{1}{2n} \sum_{i=1}^n 
    \norm{
        \left(\nabla f(\tilde{x}_i) - J^{(i)}\right)^\top 
        \left(\nabla^2_{uu}\ell(\tilde{y}_i,f(\tilde{x}_i))\right)^\frac{1}{2}
    }^2_{\Sigma_{\tilde{x}\tilde{x}}^{(i)}}\,,\\
R_2(f) &= \frac{1}{2n} \sum_{i=1}^n \scal{ \Sigma_{\tilde{x}\tilde{x}}^{(i)}
}{ \nabla_u\ell(\tilde{y}_i,f(\tilde{x}_i))\nabla^2 f(\tilde{x}_i)} \,,\\
R_3(f) &= -\frac{1}{2n} \sum_{i=1}^n
    \norm{
        \Sigma_{\tilde{x}\tilde{y}}^{(i)}\,\,
        \nabla^2_{yu}\ell(\tilde{y}_i,f(\tilde{x}_i))
        \left(\nabla^2_{uu}\ell(\tilde{y}_i,f(\tilde{x}_i))\right)^{-\frac{1}{2}}
    }^2_{\left(\Sigma_{\tilde{x}\tilde{x}}^{(i)}\right)^{-1}}\,,\\
R_4(f) &= \frac{1}{2n} \sum_{i=1}^n \scal{  \Sigma_{\tilde{y}\tilde{y}}^{(i)}
    }{\nabla^2_{yy}\ell(\tilde{y}_i,f(\tilde{x}_i))} \,,
\end{align*}
and
\begin{equation}\label{eq:Ji}
\forall i\in[n],\quad J^{(i)} = - \left(\nabla^2_{uu}\ell(\tilde{y}_i,f(\tilde{x}_i))\right)^{-1} \nabla^2_{uy}\ell(\tilde{y}_i,f(\tilde{x}_i)) \Sigma_{\tilde{y}\tilde{x}}^{(i)} \left(\Sigma_{\tilde{x}\tilde{x}}^{(i)}\right)^{-1}\,.
\end{equation}
\end{theorem}

\Cref{thm:mixupreg} captures the effect of the random perturbations in Mixup as a sum of four penalty terms $R_i(f)$ for $i\in[4]$. 
They regularize the simple ERM risk applied on the modified inputs $\tilde{x}_i$ and smoothed outputs $\tilde{y}_i$. 
\rev{Before discussing the accuracy and practical consequences of this reformulation of Mixup as regularized empirical risk minimization on modified data in the next Section, we now derive the details of this approximation for the cross-entropy, logistic and squared error losses.}
We begin by presenting the results for the cross-entropy loss:

\begin{corollary}\label{cor:CEloss}
Let $\Scal:\R^c \rightarrow \R^c$ be the softmax operator, i.e., 
for any $i\in[c]$ and $u\in\R^c$, $\Scal(u)_i =  e^{u_i} / \sum_{j=1}^c e^{u_j}$, 
and let $H(u) = \text{diag}(\Scal(u)) - \Scal(u)\Scal(u)^\top \in\R^{c\times c}$. The approximate Mixup risk for the cross-entropy loss satisfies
$$
 \Ex^{\text{Mixup}}_Q(f)= \frac{1}{n} \sum_{i=1}^n  \ell^{\text{CE}}( \tilde{y}_i , f(\tilde{x}_i)) + R_1^{\text{CE}}(f) + R_2^{\text{CE}}(f) + R_3^{\text{CE}}(f)\,,
$$
where
\begin{align*}
    R_1^{\text{CE}}(f) &= \frac{1}{2n} \sum_{i=1}^n 
    \norm{ 
        \left(\nabla f(\tilde{x}_i) - J^{(i)}\right)^\top 
        H(f(\tilde{x}_i))^{\frac{1}{2}}
    }^2_{\Sigma_{\tilde{x}\tilde{x}}^{(i)}} \,,\\
R_2^{\text{CE}}(f) &= \frac{1}{2n} \sum_{i=1}^n \scal{\Sigma_{\tilde{x}\tilde{x}}^{(i)}}{ \left(  \Scal(f(\tilde{x}_i)) - \tilde{y}_i \right)^\top \nabla^2 f(\tilde{x}_i)} \,,\\
R_3^{\text{CE}}(f) &= -\frac{1}{2n} \sum_{i=1}^n
    \norm{\Sigma_{\tilde{x}\tilde{y}}^{(i)}\,\, H(f(\tilde{x}_i))^{-\frac{1}{2}} }^2_{\left(\Sigma_{\tilde{x}\tilde{x}}^{(i)}\right)^{-1}}\,,\\\,
\end{align*}
with
\begin{equation}\label{eq:JiCE}
\forall i\in[n]\,,\quad J^{(i)} = H(f(\tilde{x}_i))^{-1} \Sigma_{\tilde{y}\tilde{x}}^{(i)} \left(\Sigma_{\tilde{x}\tilde{x}}^{(i)}\right)^{-1}\,.
\end{equation}
\end{corollary}

In the binary classification setting, minimizing the empirical cross-entropy risk over $f:\X\rightarrow\R^2$ after one-hot encoding of the two possible classes in $\R^2$ as $(0,1)^\top$ and $(1,0)^\top$ is equivalent to minimizing the following well-known logistic loss over $f:\X\rightarrow \R$ after encoding the two classes in $\R$ as $0$ and $1$:
\begin{equation}\label{eq:lrloss}
    \ell^{\text{LR}}(y, u)= \log(1 + e^u) - yu \,.
\end{equation}
The regularization effect of Mixup in that case is detailed in the following result:
\begin{corollary}\label{cor:logisticloss}
Let $s:\R \rightarrow \R$ be the sigmoid operator, i.e., 
for any $u\in\R$, $s(u) = (1 + e^{-u})^{-1}$, and 
let $v(u) = s(u)(1-s(u)) \in\R$. The approximate Mixup risk for the logistic regression loss satisfies
$$
 \Ex^{\text{Mixup}}_Q(f)= \frac{1}{n} \sum_{i=1}^n  \ell^{\text{LR}}( \tilde{y}_i , f(\tilde{x}_i)) + R_1^{\text{LR}}(f) + R_2^{\text{LR}}(f) + R_3^{\text{LR}}(f)\,,
$$
where
\begin{align*}
    R_1^{\text{LR}}(f) &= \frac{1}{2n} \sum_{i=1}^n v(f(\tilde{x}_i)) \norm{\nabla f(\tilde{x}_i) - J^{(i)}}^2_{\Sigma_{\tilde{x}\tilde{x}}^{(i)}} \,,\\
    R_2^{\text{LR}}(f) &= \frac{1}{2n} \sum_{i=1}^n \left(s(f(\tilde{x}_i))- \tilde{y}_i\right)\scal{\Sigma_{\tilde{x}\tilde{x}}^{(i)}}{ \nabla^2 f(\tilde{x}_i)} \,,\\
    R_3^{\text{LR}}(f) &= -\frac{1}{2n} \sum_{i=1}^n v(f(\tilde{x}_i))^{-1} \Sigma_{\tilde{y}\tilde{x}}^{(i)}\left(\Sigma_{\tilde{x}\tilde{x}}^{(i)}\right)^{-1} \Sigma_{\tilde{x}\tilde{y}}^{(i)}\,,\\\,.
\end{align*}
with
\begin{equation}\label{eq:JiLR}
\forall i\in[n]\\,\quad J^{(i)} = \frac{\Sigma_{\tilde{y}\tilde{x}}^{(i)} \left(\Sigma_{\tilde{x}\tilde{x}}^{(i)}\right)^{-1}}{v(f(\tilde{x}_i))}\,.
\end{equation}
\end{corollary}

The next result summarizes the form of the approximate Mixup risk in the case of the squared error loss, 
and shows in particular that Mixup has no effect for linear least-squares regression models.
\begin{corollary}\label{cor:SEloss}
The approximate Mixup risk for the squared error loss satisfies
\begin{equation}\label{eq:QmixupSE}
 \Ex^{\text{Mixup}}_Q(f)= \frac{1}{n} \sum_{i=1}^n  \ell^{\text{SE}}( \tilde{y}_i , f(\tilde{x}_i)) + R_1^{\text{SE}}(f) + R_2^{\text{SE}}(f) + C\,,
\end{equation}
where $C$ is a constant independent of $f$ and
\begin{align*}
    R_1^{\text{SE}}(f) = \frac{1}{2n} \sum_{i=1}^n \| \nabla f(\tilde{x}_i) - J^{(i)} \|^2_{\Sigma_{\tilde{x}\tilde{x}}^{(i)}}
    \quad \text{and}\quad
    R_2^{\text{SE}}(f) = \frac{1}{2n} \sum_{i=1}^n \scal{\Sigma_{\tilde{x}\tilde{x}}^{(i)}}{\left(  f(\tilde{x}_i) - \tilde{y}_i \right)^\top \nabla^2 f(\tilde{x}_i)} \,,
\end{align*}
with
\begin{equation}\label{eq:JiSE}
\forall i\in[n]\\,\quad J^{(i)} = \Sigma_{\tilde{y}\tilde{x}}^{(i)} \left(\Sigma_{\tilde{x}\tilde{x}}^{(i)}\right)^{-1}\,.
\end{equation}
In particular, when we consider linear models with intercept of the form $f_{W,b}(x)=Wx+b$ for $(W,b)\in\R^{c\times d}\times\R^{c}$, then the exact Mixup risk satisfies \begin{equation}\label{eq:mixupOLS}
 \Ex^{\text{Mixup}}(f_{W,b})= \frac{2\sigma^2 + 2\bar{\theta}^2 + (1-\bar{\theta})^2}{2n} \sum_{i=1}^n  \ell^{\text{SE}}( y_i , f_{W,\bar{b}}(x_i)) + \|b-\bar{b}\|^2 + C \,,
\end{equation}
where $C$ is a constant that does not depend on $(W,b)$ and $\bar{b} = \bar{y} - W \bar{x}$. Consequently, the linear model that minimizes $\Ex^{\text{Mixup}}$ is the standard multivariate ordinary least squares (MOLS) predictor that minimizes $\Ex^{\text{ERM}}$ on the original data, i.e., Mixup has no effect on linear least-squares regression.
\end{corollary}

 \section{Discussion and experiments}\label{sec:discussion}

Let us now discuss \rev{how our analysis relates to a recent similar study by \citet{Zhang2021HowDM}}, and empirically assess the validity of our analysis and the regularization properties of Mixup it suggests.
To support our discussion, we provide empirical results on CIFAR-10/100 and ImageNet for different networks
(LeNet, ResNet-34/50). 
For each experimental result we report mean and 95\% confidence interval using 10 repetitions (unless stated otherwise). 
All details about experiments are provided in \Cref{sec:app_exp}, together with other experiments on 
the simpler setting of learning on the two-moon dataset with random features. 

\rev{\paragraph{Comparison with related work.}
\citet{Zhang2021HowDM} independently published a similar analysis of the regularization effect of Mixup. Both works provide complementary and coherent views of the effect of Mixup on generalization and robustness, and differ in a few technical aspects that we clarify here. First, we provide an analysis valid for any loss $\ell(y, u)$, where $y$ and $u$ are respectively the predicted and true outputs, while \citet{Zhang2021HowDM} restrict their analysis to the losses of the form $\ell(y, u) = h(u) - yu$ for some function $h$. Second, while both works use a second-order Taylor expansion to approximate the loss at a Mixup example, the expansions are different since we do a Taylor expansion near the expected value of the Mixup example, while \citet[Lemma 3.1]{Zhang2021HowDM} do a Taylor expansion near a non-Mixup example. One consequence is that our Taylor approximation converges to the exact Mixup risk when the amount of Mixup reduces (i.e., when $\alpha \rightarrow +\infty$ in the $Beta(\alpha, \alpha)$ distribution of the mixing parameter), while the one used by \citet{Zhang2021HowDM} does not\footnote{\rev{More precisely, the Taylor approximation remainder in \citet[Lemma 3.1]{Zhang2021HowDM} does not vanish since it is a Taylor expansion near 0 of a random variable that follows a $Beta(\alpha+1, \alpha+1)$ distribution, i.e., that takes values close to $1$ with probability $1/2$ when $\alpha$ goes to $+\infty$.}}. Third, the generalization analysis of \citet{Zhang2021HowDM} for generalized linear models (GLM) is performed for a variant of Mixup where the mixed input is $x_i + (1/\lambda - 1) x_j$, while we focus on the standard Mixup $\lambda x_i + (1-\lambda) x_j$. A consequence of this difference is that we identify the importance of rescaling data at test time for standard Mixup (see below), while no such rescaling is needed in the variant considered by \citet{Zhang2021HowDM}. Finally, \citet{Zhang2021HowDM} explore the link between Mixup-induced regularization and adversarial robustness and generalization through Rademacher complexity analysis, which we do not explore in this work but could be done in a similar manner.}
\\

\begin{figure*}[ht]
\hfill
    \minipage{0.32\textwidth}
    \includegraphics[width=\linewidth]{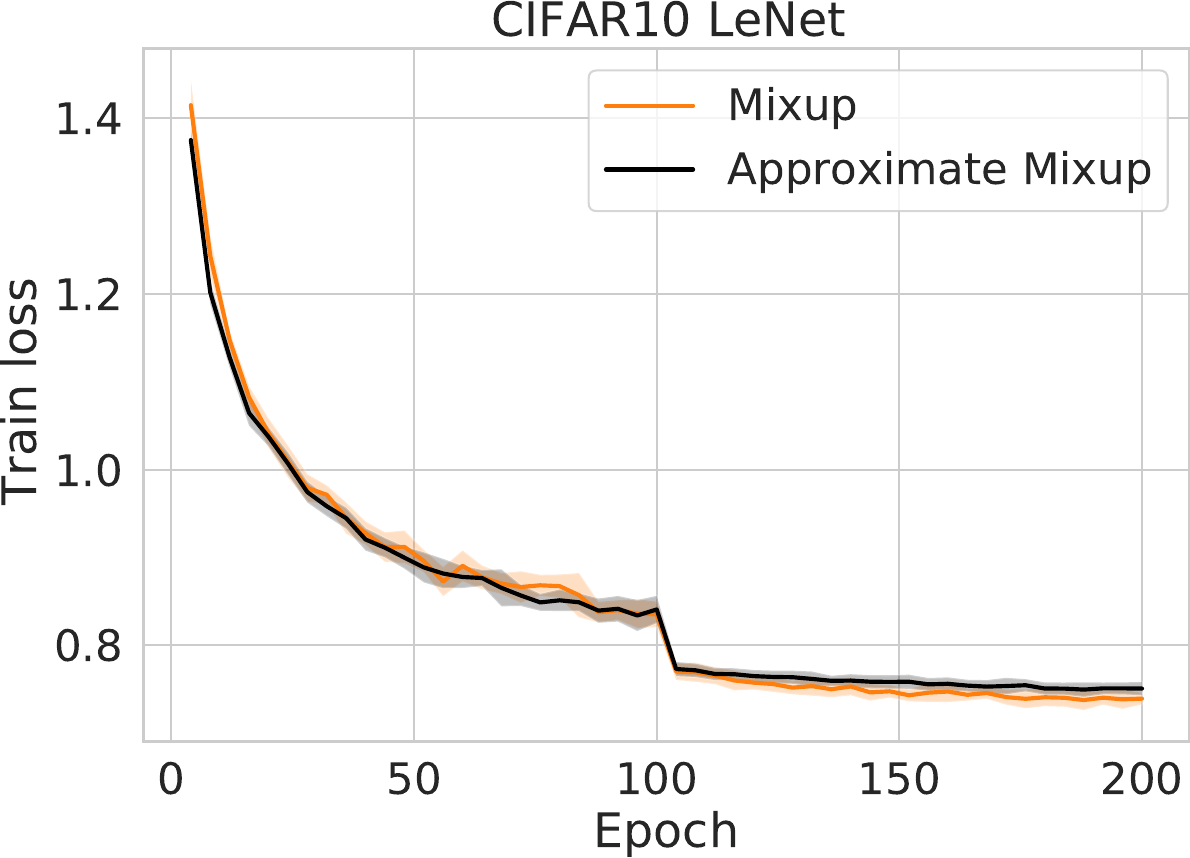}
    \endminipage\hfill
    \minipage{0.32\textwidth}
    \includegraphics[width=\linewidth]{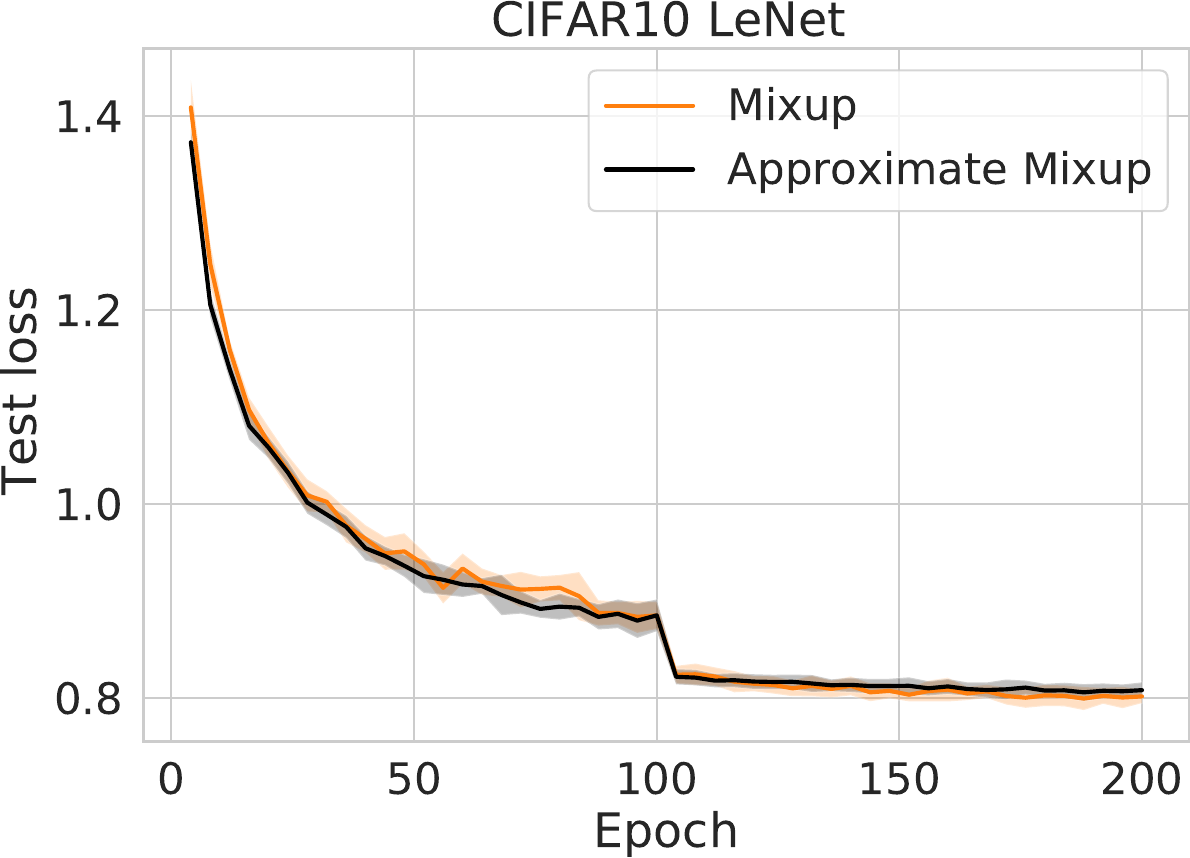}
    \endminipage\hfill 
    \minipage{0.32\textwidth}
    \includegraphics[width=\linewidth]{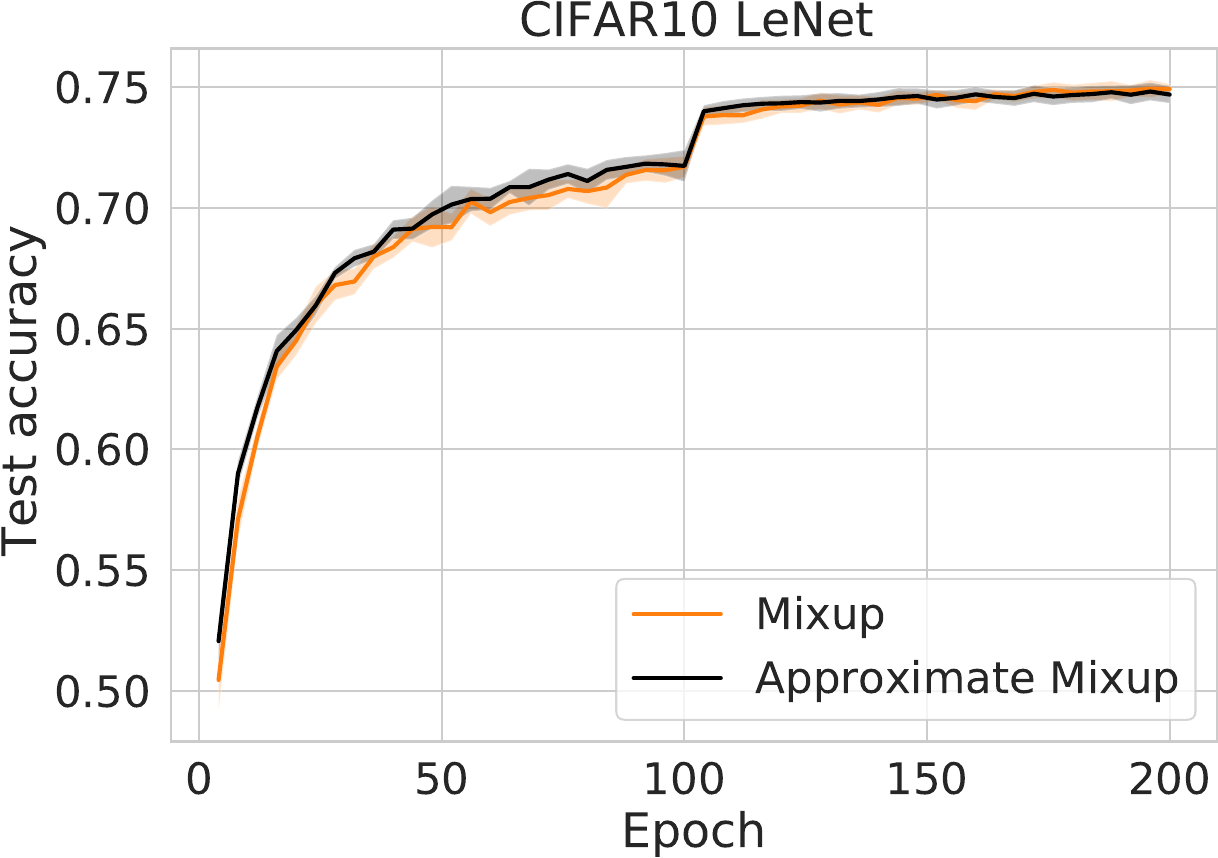}
    \endminipage\hfill
    \caption{\small{From left to right: train loss, test loss and accuracy during optimization of 
    LeNet on CIFAR-10 with Mixup and approximate Mixup.}
    }\label{fig:mess_vs_approx_acc}
\end{figure*} 
\paragraph{Validity of the Taylor approximation.} 
To analyze the regularization effect of Mixup, we used a quadratic approximation of the loss 
function (\ref{eq:taylor}). We note that compared to similar approximations that have been proposed 
to study the regularization effect of input perturbation only, such as 
dropout~\citep{wager2013dropout,wei2020implicit}, we must include in the Taylor expansion all 
second-order terms involving the input perturbation only (term with $\delta\delta^\top$), 
the output perturbation only (term with $\epsilon \epsilon^\top$), and their interaction 
(term with $\epsilon \delta^\top$). In the absence of output perturbation (e.g., in the case of dropout), 
only the term in $\delta\delta^\top$ matters, and in the absence of correlation between input and output 
perturbation (e.g., dropout combined with independent label smoothing), then the term in $\epsilon\delta^\top$ 
does not matter either. Mixup is unique in the correlation it creates between input and output perturbations, 
which is captured by the interaction term with $\epsilon\delta^\top$ in (\ref{eq:taylor}). 
Regarding the validity of the Taylor approximation, we note that, as for similar work on input perturbation, 
the approximate Mixup risk (\ref{eq:mixupQ}) is only a good approximation to the Mixup risk for ``small'' 
perturbations; as noted by \citet[Annex A.2]{wei2020implicit}, though, this often remains valid even for ``large'' 
input perturbation followed by a linear transformation layer. 
To support empirically the validity of the 
approximation, \Cref{fig:mess_vs_approx_acc} shows the training and test 
performance of training a LeNet on CIFAR-10 using
Mixup (minimizing \eqref{eq:mixup}), and using the approximate Mixup formulation (minizing \eqref{eq:mixapprox}), 
where we dropped the term $R_2(f)$ in the regularization 
since it empirically induces numerical instability due to it non-convexity \citep[see also][for a discussion about 
discarding the Hessian regularization]{wei2020implicit}. 
\begin{figure}[ht]
\centering
    \minipage{0.39\textwidth}
    \includegraphics[width=\linewidth]{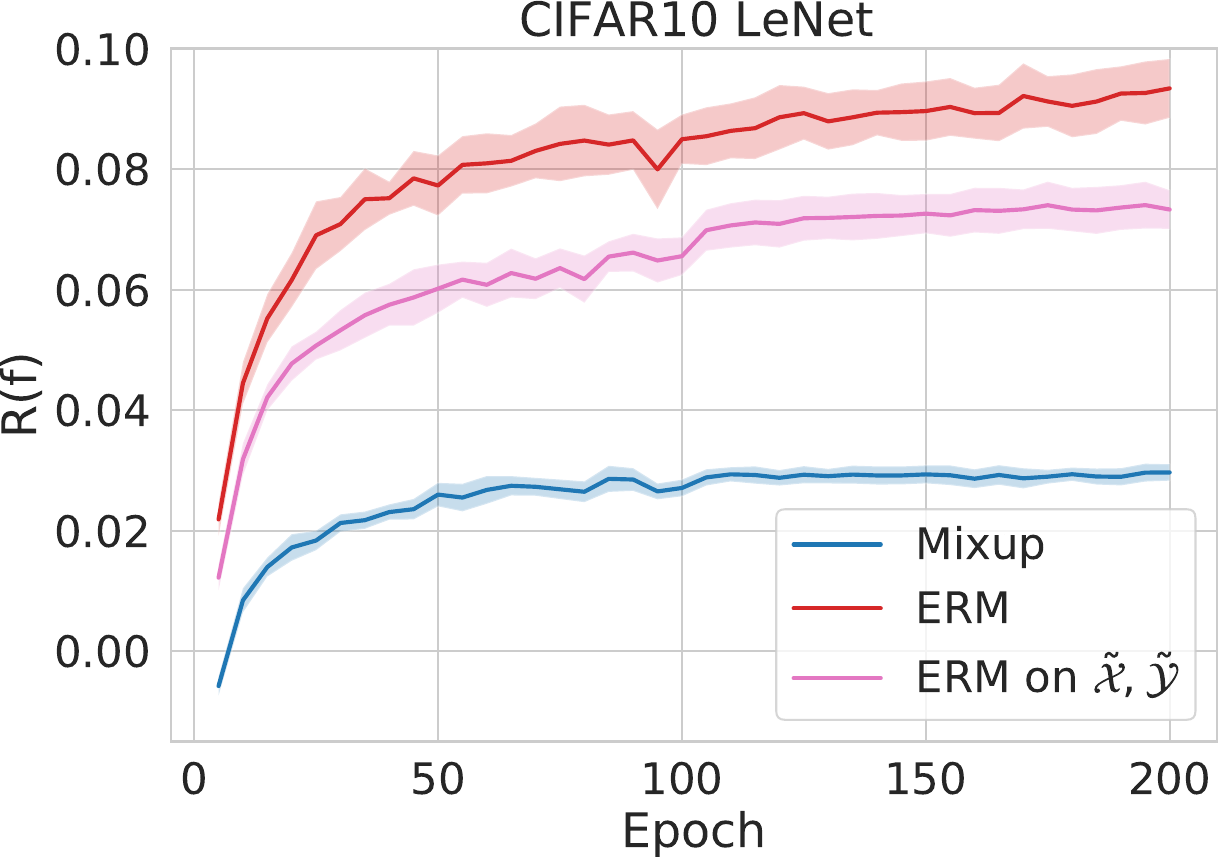}
    \endminipage
\hspace{1cm}
    \minipage{0.39\textwidth}
    \includegraphics[width=\linewidth]{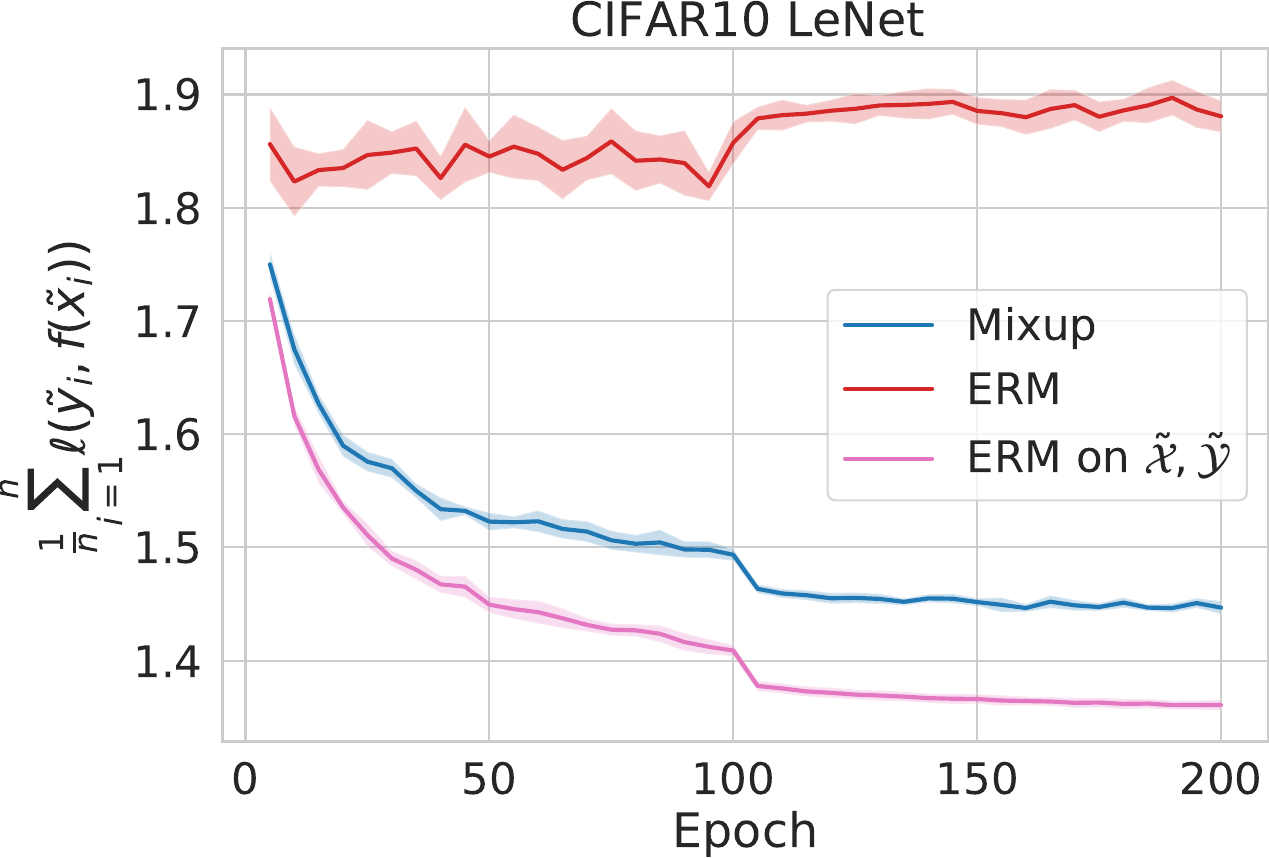}
    \endminipage
\caption{\small{Evaluations of the regularization terms of the mixup approximation (left) and of the loss on modified data (right) for functions learned with Mixup, ERM and ERM on modified data.}}\label{fig:regularization_taylor}
\end{figure}
We can see how when training using the approximate Mixup formulation, 
we learn functions which mimic, both in training and test, 
the performance of functions learned when training with the standard Mixup procedure.
To further mark the validity of the approximation and decouple the contributions of 
the data transformation \eqref{eq:modif} and the input perturbation \eqref{eq:perturbations},
we evaluate for functions learned with Mixup \eqref{eq:mixup}, ERM \eqref{eq:erm} and ERM on modified data: 
\begin{equation}\label{eq:ermtild}
    \min_{f\in\Hcal}\frac{1}{n}\sum_{i=1}^n \ell(\widetilde y_i, f(\widetilde x_i))\,,
\end{equation}
the regularization terms of the approximation \eqref{eq:mixapprox} and the loss on modified data \eqref{eq:ermtild}.
From \Cref{fig:regularization_taylor} we see that the functions learned with Mixup are the ones with the smallest values
of the regularization terms but not the smallest loss on modified data, which confirms that the model trained with Mixup finds a trade-off between empirical risk and the regularization we study.

\paragraph{Data modification at test time.} 
By \Cref{thm:mixupAsErm}, we see that Mixup implicitly shrinks inputs towards their 
mean since the Mixup risk involves the empirical risk over modified inputs $\tilde{x}_i$ and outputs $\tilde{y}_i$. 
In particular, this means that the functions that the standard Mixup procedure learns are not functions
from the space of input points $\X$ to the output points $\Y$, but it learns functions from $\widetilde \X$ to $\widetilde Y$,
which are spaces defined by the training points and the hyperparameter $\alpha$ of Mixup as  
\begin{equation}
    \begin{cases}
        \widetilde \X = \{\bar x + \bar\theta(x - \bar x)\, |\, \forall x \in \X\}\,, \\
        \widetilde \Y = \{\bar y + \bar\theta(y - \bar y)\, |\, \forall y \in \Y\}\,,
    \end{cases}
\end{equation}
where $\bar x$ and $\bar y$ are respectively the average training inputs and outputs points, 
and $\bar\theta = \E_\theta \theta$ as defined in \Cref{thm:mixupAsErm}.
An important consequence is that the function $f: \widetilde \X \rightarrow \widetilde \Y$ estimated by Mixup, 
should ideally be applied at test time to transformed data, to 
map the test point $x_{\text{test}}\in X$ to $\tilde x_{\text{test}} \in\widetilde X$, 
and the evaluation of the function $f(\tilde x_{\text{test}}) \in \widetilde Y$ should be mapped back to $\Y$.
In details, the prediction for point $x_\text{test}$ should be
\begin{equation}\label{eq:testmodif}
\rev{\text{pred}_f(x_\text{test}) = \bar{y} \left( 1 -  1 / \bar{\theta} \right) + 1 / \bar{\theta} \cdot f\left( \bar{\theta} x_\text{test} + (1-\bar{\theta} )\bar{x} \right)\,.}
\end{equation}

\begin{small}
    \begin{algorithm}[H]\small
        \caption{{\tt python} code to evaluate according to \eqref{eq:testmodif} functions learned with mixup
            \label{algo:pred-python}}
\vspace{-0.35cm}
        \begin{flushleft}
            {\bf Input:}  \texttt{X\_test} \textit{Tensor} of test points to evaluate, 
            \texttt{trained\_mode} \textit{Model} trained using mixup,
            \texttt{alpha} \textit{float} hyperparameter used by mixup during training, 
            \texttt{X\_train}, \texttt{Y\_train} \textit{Tensor} of points used for training with one-hot encoded labels.\\
        \end{flushleft}
        
     {\footnotesize
        \begin{center}
\begin{verbatim}
    import scipy.special as sc
    import torch
    
    X_bar = torch.mean(X_train, dim=0, keepdim=True)
    Y_bar = torch.mean(Y_train, dim=0, keepdim=True)
    
    # expectation of a truncated beta distribution in [0.5, 1]
    theta_bar = 1. - sc.betainc(alpha + 1., alpha, .5) 
    
    def predict_mixup(X_test, trained_model):
        f_X = trained_model.forward((1. - theta_bar) * X_bar + theta_bar * X_test)
        return Y_bar * (1. - 1. / theta_bar) + f_X / theta_bar
            \end{verbatim}
        \end{center}
    }
    \vspace{-0.35cm}
    
    \end{algorithm}
    \end{small}
For centered training data ($\bar{x}=\bar{y}=0$) and homogeneous functions 
($f(ux)=uf(x)$ for any $(u,x)\in\R^+\times\X$, e.g., linear models 
or neural networks with ReLU activation and linear transformations), 
this has no impact as $\text{pred}_f(x)=f(x)$ in that case. 
For more general models, however,  (\ref{eq:testmodif}) may be a better predictor than $f$. 
For example, we clearly see in \Cref{fig:overview} that the asymptotically Bayes optimal classifier 
under the Mixup distribution matches the one under the empirical distribution of the modified data 
(up to regularization effects), and not of the original data. 
Interestingly, when the classes are balanced, 
i.e., $\bar{y}= \frac{1}{c} \mathbf{1}_c$, the transformation in~(\ref{eq:testmodif}) adds the same 
constant to each of the $c$ entries of $f$. In particular, in the multi-class setting, 
since the softmax is invariant to a constant in the logits, (\ref{eq:testmodif}) becomes equivalent 
to a scaling of the logits, commonly referred to as temperature scaling~\citep{guo2017calibration}. 
While temperature scaling is traditionally tuned with a validation set~\citep{guo2017calibration}, 
mixup automatically sets this value, according to the distribution of $\theta$.
\begin{figure}[ht]
    \includegraphics[width=\linewidth]{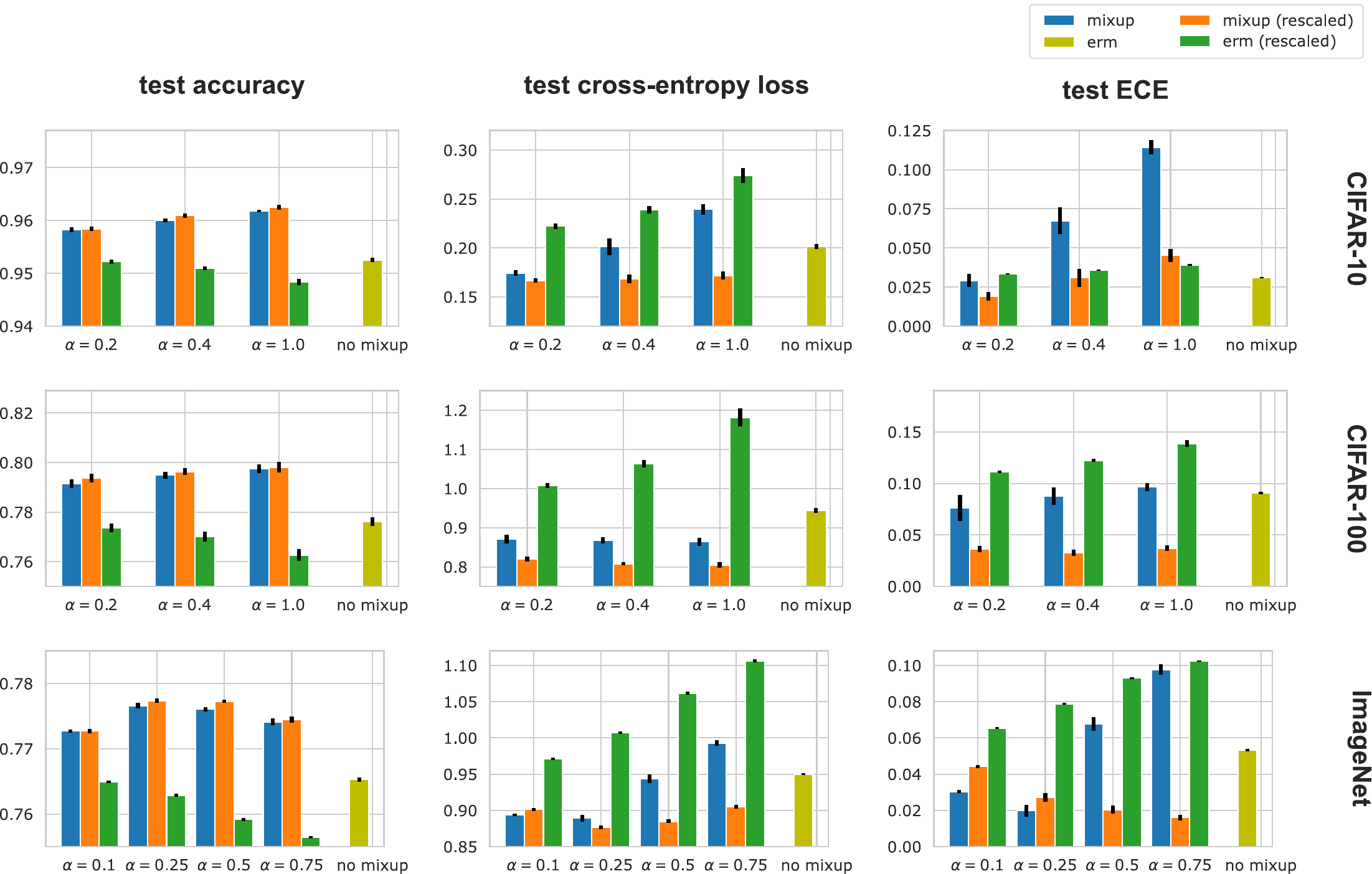}
    \caption{\small{Test accuracy, test cross-entropy loss and test expected calibration error (first, second and third column respectively) on the CIFAR-10, CIFAR-100, ImageNet datasets (first, second and third row respectively). 
    We report the mean and the standard error over 10 repetitions for CIFAR-10 and CIFAR-100 and 5 repetitions for ImageNet. 
    The training was done with ResNet-34 for CIFAR-10, CIFAR-100, and with ResNet-50 for ImageNet.}
    }\label{fig:data_transf_mixup}
\end{figure} 
To point out the advantages of using \eqref{eq:testmodif}, we compare in \Cref{fig:data_transf_mixup} the performance
of ERM, standard Mixup (for different $\alpha$ values), the same Mixup but with the proper 
data transformation \eqref{eq:testmodif} at test time and the same data transformation applied to the ERM estimator. 
The trained networks are ResNet-34 for CIFAR-10 and CIFAR-100, and ResNet-50 for ImageNet.
For CIFAR-10 and 100, we observe overall benefits of using the data transformation for evaluating the functions learned with Mixup:
higher test accuracy, lower test loss, and lower expected calibration error (ECE). For ImageNet we have the same benefits, 
with the only exception of the test loss and the ECE for very low values of $\alpha$. 
Notice indeed that for small values of $\alpha$ the data transformation \eqref{eq:testmodif} has a smaller impact
than for bigger values of $\alpha$, as $\lim_{\alpha \rightarrow 0} \bar\theta = 1$ 
while $\lim_{\alpha \rightarrow +\infty} \bar\theta = \frac{1}{2}$. Thus, as observed empirically, the 
``correction'' \eqref{eq:testmodif} is more important, as it brings bigger improvements, 
when training Mixup with bigger $\alpha$. Finally, we can observe that when 
the data transformation \eqref{eq:testmodif} is applied to ERM, performance always deteriorate.
This supports that \eqref{eq:testmodif} is a mixup specific improvement. 
\Cref{algo:pred-python} shows the few lines of codes that implement the new prediction procedure \eqref{eq:testmodif}.
\begin{figure}[ht]
    \includegraphics[width=\linewidth]{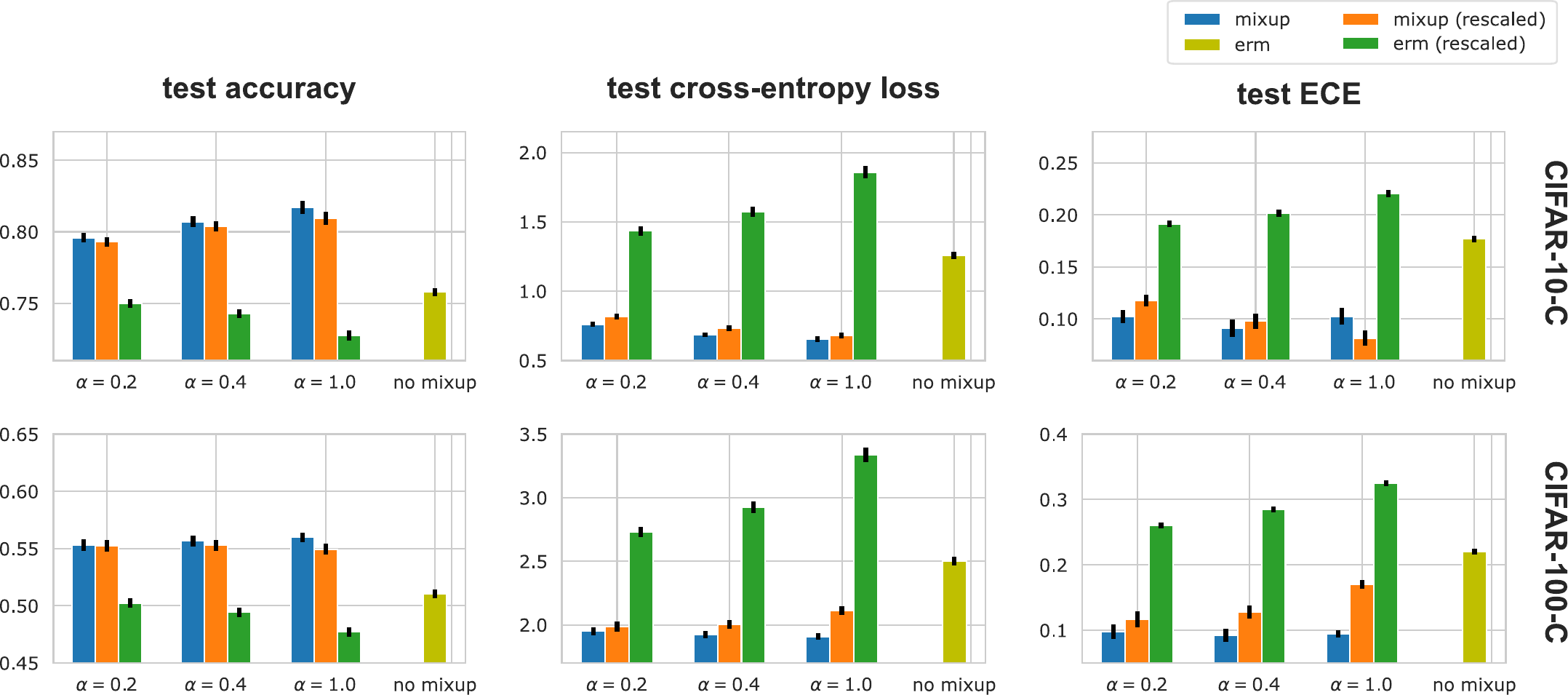}
    \caption{\small{Test accuracy, test cross-entropy loss and test expected calibration error (first, second and third column respectively) 
    on the CIFAR-10-C, CIFAR-100-C (first, and second row respectively). 
    We report the mean and the standard error over 10 repetitions.
    The training was done with ResNet-34 on the standard CIFAR-10, CIFAR-100.}
    }\label{fig:data_transf_mixup-OOD}
\end{figure}
\paragraph{Data modification for out-of-distribution data.} 
Even though our theoretical analysis holds only for in-distribution data, we now empirically investigate whether the benefits of rescaling data at test time that we observe for models trained with Mixup also hold when the test time data come from a different distribution, a setting called out-of-distribution (OOD) data \citep{hendrycks2016baseline}. Indeed, standard Mixup is known to provide some benefits in out-of-distribution settings compared to models simply trained by ERM, so it is interesting to assess whether the data modification scheme we propose can further boost the accuracy and calibration of models trained with Mixup in that setting.
In particular we consider CIFAR-10-C, CIFAR-100-C, ImageNet-C \citep{hendrycks2019robustness}, ImageNet-A \citep{hendrycks2019natural}, 
ImageNetV2 \citep{recht2019imagenet}, ImageNet-Vid-Robust, YTBB-Robust \citep{shankar2019image}, 
ObjectNet \citep{barbu2019objectnet}. 
These benchmark datasets are designed by systematically perturbing in a controlled way 
the in-distribution data (CIFAR-10, CIFAR-100 and ImageNet), e.g., 
by adding noise or applying transformations to the images.
\begin{figure}[ht]
    \includegraphics[width=\linewidth]{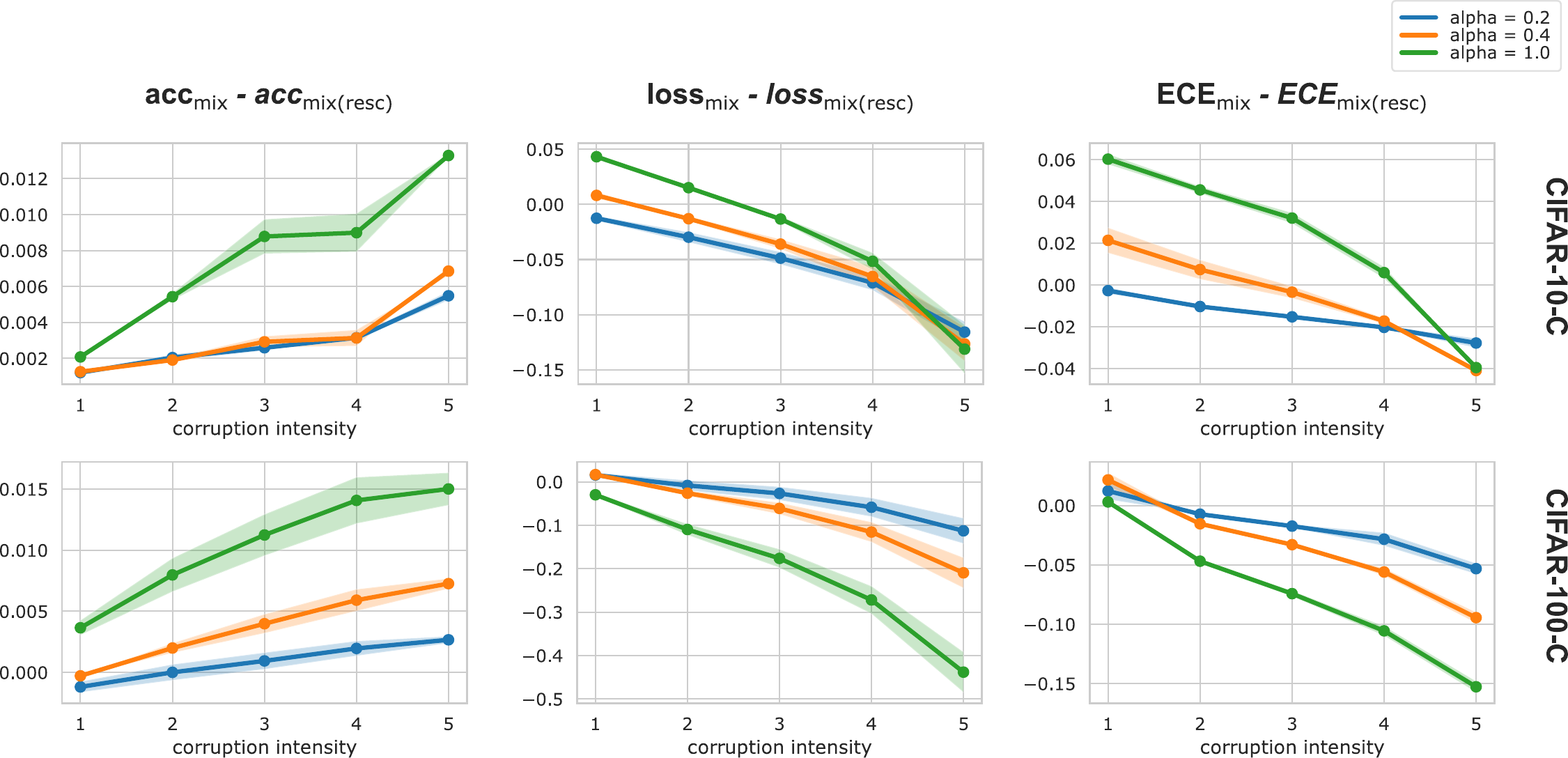}
    \caption{\small{Difference in test performance between Mixup and rescaled Mixup on CIFAR-10-C, CIFAR-100-C for the different corruption intensities. 
    The training was done with ResNet-34 on the standard CIFAR-10, CIFAR-100 dataset and the performance metrics are accuracy, cross-entropy loss, expected calibration error.}}\label{fig:diff-rescaling_mixup}
\end{figure} 
\Cref{tab:imagenet-ood} details the performance of models trained with Mixup for $\alpha=0.25$ and $\alpha=0.5$, with or without data modification at test time, on out-of-distribution datasets derived from ImageNet, while \Cref{fig:data_transf_mixup-OOD} shows similar results for benchmarks derived from CIFAR-10 and CIFAR-100. We see that for most datasets the rescaling improves accuracy and some of the other metrics, 
but for others and in particular CIFAR-10-C, CIFAR-100-C and ImageNet-C, the rescaling worsen almost all metrics with respect of the standard Mixup.
We now investigate how the performance of Mixup with or without rescaling differs with respect to the intensity of the noise.
\begin{table}[H]
    \begin{center}
    \begin{small}
    \setlength\tabcolsep{4pt}
    \begin{tabular}{l l c c c c }
        \toprule    
        \multirow{2}{*}{ImageNet} & \multirow{2}{*}{Metric} &  mixup(rescaled) &  mixup &  mixup(rescaled) &  mixup  \\
         \cmidrule(lr){3-4} \cmidrule(lr){5-6} 
        {} & {} &  \multicolumn{2}{c}{$\alpha = 0.25$}  &  \multicolumn{2}{c}{$\alpha = 0.5$} \\
        \midrule
        \multirow{3}{*}{standard} & Acc                 &                      $\bf 0.7764 {\scriptstyle\pm 0.0022}$ &                        $0.7754 {\scriptstyle\pm 0.0021}$ &                       $\bf 0.7780 {\scriptstyle\pm 0.0006}$ &                       $0.7766 {\scriptstyle\pm 0.0009}$\\
        & CE-loss                      &                      $\bf 0.8820 {\scriptstyle\pm 0.0092}$ &                        $0.8964 {\scriptstyle\pm 0.0113}$ &                       $\bf 0.8852 {\scriptstyle\pm 0.0046}$ &                       $0.9482 {\scriptstyle\pm 0.0188}$\\
        & ECE                      &                      $0.0291 {\scriptstyle\pm 0.0031}$ &                        $\bf 0.0236 {\scriptstyle\pm 0.0058}$ &                       $\bf 0.0280 {\scriptstyle\pm 0.0040}$ &                       $0.0707 {\scriptstyle\pm 0.0127}$\\
        \midrule
        \multirow{3}{*}{A} & Acc               &                      $\bf 0.0165 {\scriptstyle\pm 0.0011}$ &                        $\bf 0.0165 {\scriptstyle\pm 0.0011}$ &                       $\bf 0.0229 {\scriptstyle\pm 0.0004}$ &                       $0.0225 {\scriptstyle\pm 0.0005}$\\
        & CE-loss                    &                      $7.3395 {\scriptstyle\pm 0.0884}$ &                        $\bf 6.8075 {\scriptstyle\pm 0.0687}$ &                       $7.1010 {\scriptstyle\pm 0.0879}$ &                       $\bf 6.3738 {\scriptstyle\pm 0.0506}$\\
        & ECE                    &                      $0.4322 {\scriptstyle\pm 0.0072}$ &                        $\bf 0.3629 {\scriptstyle\pm 0.0075}$ &                       $0.4318 {\scriptstyle\pm 0.0110}$ &                       $\bf 0.3161 {\scriptstyle\pm 0.0117}$\\
        \midrule
        \multirow{4}{*}{C} & Acc          &                      $0.4638 {\scriptstyle\pm 0.0021}$ &                        $\bf 0.4662 {\scriptstyle\pm 0.0019}$ &                       $0.4719 {\scriptstyle\pm 0.0015}$ &                       $\bf 0.4761 {\scriptstyle\pm 0.0011}$\\
        & CE-loss               &                      $2.8050 {\scriptstyle\pm 0.0140}$ &                        $\bf 2.7374 {\scriptstyle\pm 0.0094}$ &                       $2.7842 {\scriptstyle\pm 0.0164}$ &                       $\bf 2.7206 {\scriptstyle\pm 0.0136}$\\
        & MCE                   &                      $0.6824 {\scriptstyle\pm 0.0025}$ &                        $\bf 0.6795 {\scriptstyle\pm 0.0022}$ &                       $0.6735 {\scriptstyle\pm 0.0018}$ &                       $\bf 0.6685 {\scriptstyle\pm 0.0014}$\\
        & ECE               &                      $0.1040 {\scriptstyle\pm 0.0046}$ &                        $\bf 0.0652 {\scriptstyle\pm 0.0019}$ &                       $0.1096 {\scriptstyle\pm 0.0054}$ &                       $\bf 0.0793 {\scriptstyle\pm 0.0061}$\\
        \midrule
        \multirow{3}{*}{V2} & Acc              &                      $\bf 0.6560 {\scriptstyle\pm 0.0028}$ &                        $0.6543 {\scriptstyle\pm 0.0029}$ &                       $\bf 0.6593 {\scriptstyle\pm 0.0007}$ &                       $0.6575 {\scriptstyle\pm 0.0011}$\\
        & CE-loss                   &                      $1.5208 {\scriptstyle\pm 0.0118}$ &                        $\bf 1.5048 {\scriptstyle\pm 0.0122}$ &                       $\bf 1.5130 {\scriptstyle\pm 0.0057}$ &                       $1.5295 {\scriptstyle\pm 0.0170}$\\
        & ECE                   &                      $0.0746 {\scriptstyle\pm 0.0053}$ &                        $\bf 0.0344 {\scriptstyle\pm 0.0030}$ &                       $0.0715 {\scriptstyle\pm 0.0079}$ &                       $\bf 0.0417 {\scriptstyle\pm 0.0096}$\\
        \midrule
        Vid-Robust & Acc (pm-k)  &                      $\bf 0.5254 {\scriptstyle\pm 0.0043}$ &                        $0.5244 {\scriptstyle\pm 0.0036}$ &                       $\bf 0.5224 {\scriptstyle\pm 0.0023}$ &                       $0.5216 {\scriptstyle\pm 0.0035}$\\
        \midrule
        YTBB-Robust & Acc (pm-k)          &                      $\bf 0.4738 {\scriptstyle\pm 0.0037}$ &                        $0.4717 {\scriptstyle\pm 0.0035}$ &                       $\bf 0.4705 {\scriptstyle\pm 0.0021}$ &                       $0.4692 {\scriptstyle\pm 0.0017}$\\
        \midrule
        ObjectNet & Acc &                      $\bf 0.2756 {\scriptstyle\pm 0.0035}$ &                        $0.2751 {\scriptstyle\pm 0.0034}$ &                       $\bf 0.2837 {\scriptstyle\pm 0.0010}$ &                       $0.2825 {\scriptstyle\pm 0.0011}$\\
        \bottomrule
    \end{tabular}
    \end{small}
    \end{center}
    \caption{\small Test performance of Mixup and rescaled Mixup on different OOD versions of ImageNet.
    The training was done with ResNet-50 on the standard ImageNet dataset and the performance metrics are accuracy, cross-entropy loss, expected calibration error and mean corruption error \citep{hendrycks2019robustness}.
    For each metric means and standard errors over 10 repetitions are reported.}
    \label{tab:imagenet-ood}
\end{table}
In details, the CIFAR-10-C and CIFAR-100-C data are 95 different corrupted versions each of the original CIFAR-10, CIFAR-100 test sets: 
19 different corruption types, for 5 different growing corruption intensities. 
The results that we reported in \Cref{fig:data_transf_mixup-OOD} are the averages of the performance of 
these 95 test sets.
In \Cref{fig:diff-rescaling_mixup} instead, for Mixup and rescaled Mixup, we compute the average of the different metrics 
(test accuracy, test cross-entropy loss, ECE) over the 19 corruption types for each corruption intensity, and we 
report the difference in performance between Mixup and rescaled Mixup.
We can see that as the noise intensity grows the gap between the two methods always grows in favor of standard Mixup, 
and that only for a few metrics and settings with low noise the rescaled version is better than the standard one.
This observations encourage future investigations of the effect of Mixup on OOD data\footnote{Codes for reproducing results for ImageNet and various OOD variants are available at \url{https://github.com/google/uncertainty-baselines/tree/master/baselines/imagenet}}.

\begin{figure*}[ht]
\centering
    \includegraphics[width=0.49\textwidth]{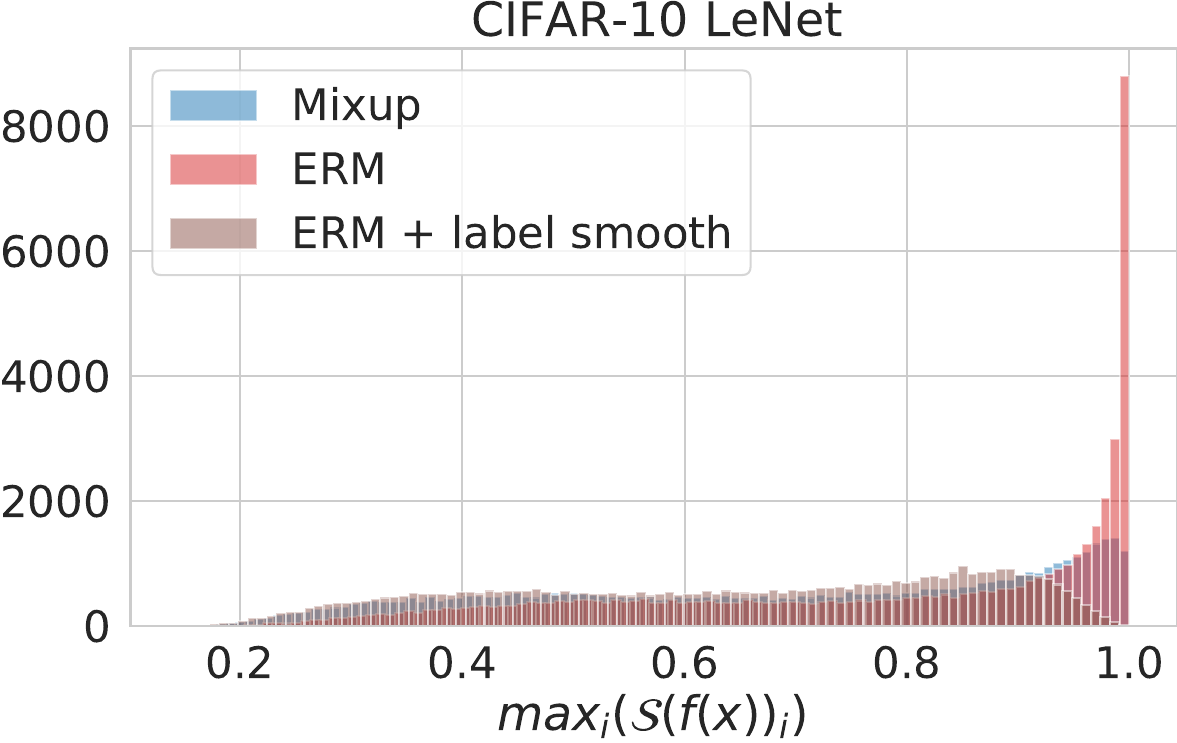}
    \includegraphics[width=0.49\linewidth]{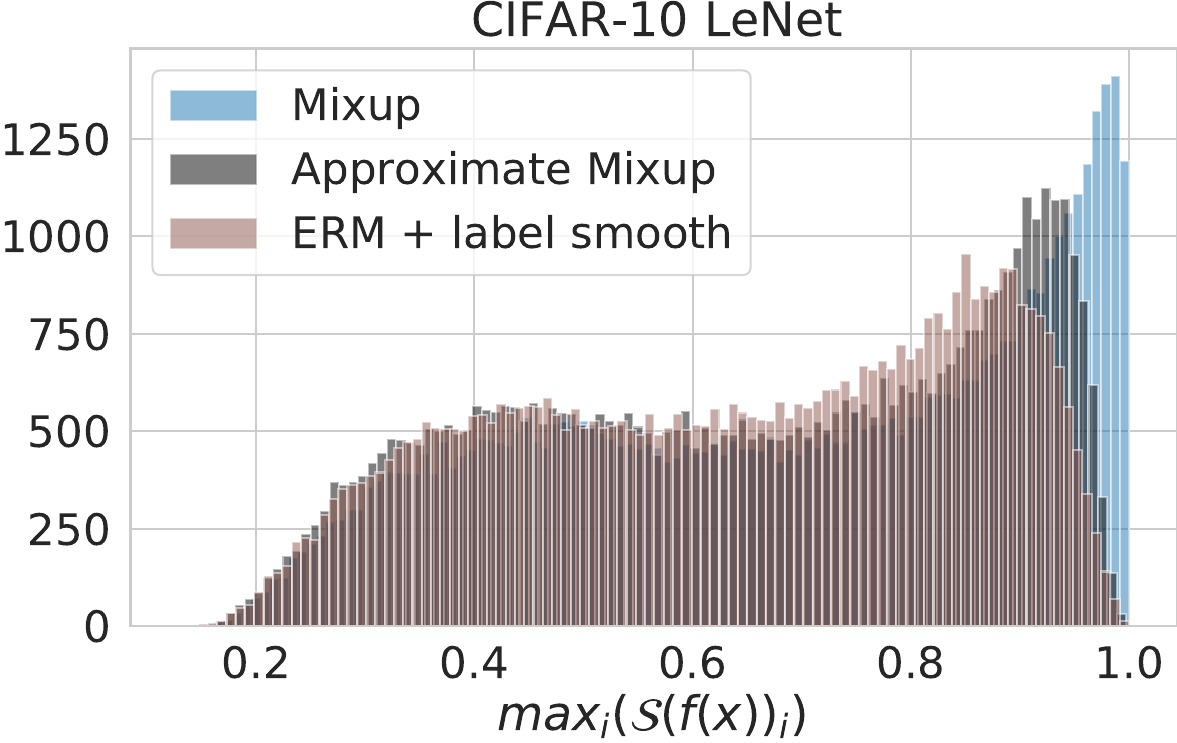}
    \caption{
        \small{Histograms of confidence of predictions for models trained with different techniques.}
        }\label{fig:confidence-lenet}
\end{figure*}

\paragraph{Label smoothing.}
The transformation that modifies the original labels $y_i$ onto $\tilde{y}_i$ acts as some form of \textit{label smoothing}, a technique known to often improve accuracy and calibration~\citep{szegedy2016rethinking,Muller2019When}. The transformed labels $\tilde{y}_i$ are indeed pulled towards the average label $\bar{y}$. Recall from \citet{szegedy2016rethinking}
that label smoothing consists in training a model on the perturbed version of the training labels defined as $y_i^{LS} = (1-\eps) y_i + \eps u(i)$, where $\eps$ is a fixed
scalar in $[0,1]$ and $u(i)$ is a fixed distribution over the labels. 
It is easy to see that for $\eps = (1 - \bar \theta)$ and $u(i) = \bar y$ the two formulations coincide. 
This implies that Mixup implicitly performs label smoothing, and can benefit from this technique in terms of accuracy or calibration.
In the following Proposition~\ref{prop:label_smoothing_logistic_reg}, we formally prove that, 
in the case of the cross entropy and linear models, 
label smoothing translates into an \textit{increase} in the average entropy of the predictions, or, 
in other words, that predictions become less certain, as observed in practice. Like in~\Cref{cor:CEloss}, we use the softmax operator $\Scal: \R^c \mapsto \Delta_c$ defined for $j \in [c]$ by $\Scal(u)_j = e^{u_j} / \left( \sum_{k=1}^c e^{u_k} \right)$:
\begin{proposition}\label{prop:label_smoothing_logistic_reg}
Let us consider the following two classification problems with a cross-entropy loss and linear model $f(x) = W x$ parameterized by $W\in\R^{c\times d},$ 
\begin{equation}\label{eq:primal_logistic_reg_without_ls}
\min_{W\in\R^{c\times d}} \frac{1}{n} \sum_{i=1}^n  \ell^{\text{CE}}( {y}_i , W x_i)
\end{equation}
and
\begin{equation}\label{eq:primal_logistic_reg_with_ls}
\min_{W\in\R^{c\times d}} \frac{1}{n} \sum_{i=1}^n  \ell^{\text{CE}}( \tilde{y}_i , W x_i)
\end{equation}
defined without and with label smoothing respectively, i.e., with $\tilde{y}_i = \bar{y} + \bar{\theta} (y_i - \bar{y} ) \in \Delta_c$ for $i \in [n]$. Let us denote by $W^\star$ and $W_\text{ls}^\star$ a solution of~(\ref{eq:primal_logistic_reg_without_ls}) and (\ref{eq:primal_logistic_reg_with_ls}) respectively, together with 
\begin{equation}\label{eq:prediction_softmax_without_with_ls}
    p_i = \Scal(W^\star x_i)
\quad \text{and}\quad
    \tilde{p}_i =\Scal(W_\text{ls}^\star x_i).
\end{equation}
It holds that the average entropy of the predictions of $W_\text{ls}^\star$ is lower bounded as follows
\begin{equation*}\bar{\theta}\frac{1}{n} \sum_{i=1}^n \Hcal(p_i) +  (1-\bar{\theta})\Hcal(\bar{y})
\leq 
\frac{1}{n} \sum_{i=1}^n \Hcal( \tilde{p}_i).
\end{equation*}
If predicting with $W^\star$ also reduces the entropy of the average predictor, i.e., $\frac{1}{n} \sum_{i=1}^n \Hcal(p_i) \leq \Hcal(\bar{y})$, then label smoothing increases the average entropy of the predictions:
\begin{equation*}
\frac{1}{n} \sum_{i=1}^n \Hcal(p_i)
\leq 
\frac{1}{n} \sum_{i=1}^n \Hcal( \tilde{p}_i).
\end{equation*}
\end{proposition}
To illustrate how both Mixup and label smoothing increase the entropy of predictions compared to ERM, we show on \Cref{fig:confidence-lenet} the histograms of 
the confidence of the estimators' predictions on test points, for a LeNet neural network trained on CIFAR-10. 
From the first plot, we notice how standard ERM produces very confident predictions, 
how label smoothing helps decreasing ERM confidence at test time, 
and how Mixup naturally produces even less confident predictions. 
From the second plot, we see that that approximate Mixup, like Mixup, produces less confident prediction, which confirms that the Mixup approximation we study captures well this behavior of Mixup.

\paragraph{Jacobian regularization.} The first implicit regularization term $R_1(f)$ in \Cref{thm:mixupreg} penalizes the discrepancy between $\nabla f(\tilde{x}_i)$ and $J^{(i)}$ given by (\ref{eq:Ji}). We recognize in $J^{(i)}$ the Jacobian of the standard MOLS model trained in the input space on the modified training set, with an increased weight for sample $(\tilde{x}_i,\tilde{y}_i)$ in $J^{(i)}$
. Compared to, e.g., dropout regularization with penalizes the norm of $\nabla f$ at the training points, we therefore see that Mixup also regularizes the Jacobian of $f$ but with a different and more informative implicit bias, namely, to mimic a good linear model in the input space. Furthermore, we note from the proof of \Cref{thm:mixupreg} that this implicit bias results from the correlation between input and output noise, which may explain why independent Mixup in the input and output performs more poorly than standard Mixup~\citep{Zhang2018mixup}. While this regularization is similar across all points in the squared loss setting (Corollary~\ref{cor:SEloss}), it is weighted by the Hessian $H(p(f(\tilde{x}_i)))$ in the cross-entropy loss (Corollary~\ref{cor:CEloss}). Similar to dropout, this implies that this regularization vanishes when the prediction $p(f(\tilde{x}_i))$ is confidently near $0$ or $1$. In the Mixup case, though, the label smoothing effect discussed in the previous paragraph tends to prevent over-confident predictions on the training point (see Proposition~\ref{prop:label_smoothing_logistic_reg} for a formal description of that property), therefore ensuring that the Jacobian regularization in $R_1(f)$ remains active even for ``easy'' points. This interaction between label smoothing (due to output Mixup) and Jacobian regularization (due to input Mixup) may explain why Mixup on inputs only performs poorly compared to Mixup on both inputs and outputs \citep{thulasidasan2019mixup}.

\section{Conclusions}
In this paper we have proposed the first theoretical analysis that explains the multiple regularization effects of Mixup. We have proved that training with Mixup is equivalent to learning on modified data with the injection of structured noise.
Through a Taylor approximation, we have further shown that 
Mixup
amounts to
empirical risk minimization on modified points plus multiple regularization terms. Fascinatingly,
our derivation shows that Mixup induces varied and complex effects, e.g., calibration, Jacobian regularization, label noise and normalization, 
while being a simple and cheap data augmentation technique.
Further, we have shown how this analysis points out a missing rescaling procedure required to evaluate functions learned with
Mixup, and we brought empirical evidence that implementing it improves accuracy and calibration.
More broadly, we have studied how a specific combination of data modification and noise injection leads to certain regularizers. An interesting research question is whether we can \textit{reverse-engineer} this process, namely starting from possibly expensive regularizers and design the corresponding data augmentation technique emulating their effects at a lower computational cost.

\section*{\bf Acknowledgments\ \ }
{The authors thank Jeremiah Zhe Liu for his feedback on an early version of this work.}

\bibliography{biblio}
 \newpage
\appendix
\section*{Appendix}
\section{Proofs}\label{sec:annex}

\subsection{Proof of \Cref{thm:mixupAsErm}}

\begin{proof}
To simplify notations, let us denote, for any $i,j\in[n]$ and $u\in[0,1]$,
$$
m_{ij}(u) = \ell\left(u y_i + (1-u)y_j , f(u x_i + (1-u) x_j)\right)\,.
$$
The Mixup risk (\ref{eq:mixup}) can then be written as
\begin{equation}\label{eq:mixupshort}
\Ex^{\text{Mixup}}(f)= \frac{1}{n^2}\sum_{i=1}^n \sum_{j=1}^n \E_\lambda m_{ij}(\lambda)\,,\quad \lambda \sim \text{Beta}(\alpha,\alpha)\,.
\end{equation}
We now separate the values of $\lambda$ depending on whether or not they are above $1/2$ by expressing it as
\begin{equation}\label{eq:betatrick}
\lambda = \pi \lambda_0 + (1-\pi)\lambda_1\,,\quad \lambda_0\sim\text{Beta}_{[0,\frac{1}{2}]}(\alpha,\alpha)\,,\quad \lambda_1\sim\text{Beta}_{[\frac{1}{2},1]}(\alpha,\alpha)\,,\quad\pi\sim\text{Ber}\left(\frac{1}{2}\right)\,.
\end{equation}
By symmetry of the $\text{Beta}(\alpha,\alpha)$ distribution around $1/2$, it is clear that $\lambda$ defined in (\ref{eq:betatrick}) follows a $\text{Beta}(\alpha,\alpha)$ distribution, and furthermore that $\lambda_1' = 1-\lambda_0$ follows, like $\lambda_1$, a $\text{Beta}_{[\frac{1}{2},1]}(\alpha,\alpha)$ distribution. For any $i,j\in[n]$, we therefore get
\begin{equation*}
    \begin{split}
        \E_\lambda m_{ij}(\lambda) 
        &= \E_{\lambda_0, \lambda_1, \pi} m_{ij}(\pi \lambda_0 + (1-\pi)\lambda_1) \\
        &= \frac{1}{2}\left[\E_{\lambda_0} m_{ij}(\lambda_0) + \E_{\lambda_1} m_{ij}(\lambda_1) \right] \\
        &= \frac{1}{2}\left[\E_{\lambda'_1} m_{ji}(\lambda'_1) + \E_{\lambda_1} m_{ij}(\lambda_1) \right]\,,
    \end{split}
\end{equation*}
where we used the fact that $m_{ij}(1-u)=m_{ji}(u)$ to get the third equality. Plugging this equality back into (\ref{eq:mixupshort}), we finally get
\begin{equation}\label{eq:technical}
    \begin{split}
        \Ex^{\text{Mixup}}(f)
        &= \frac{1}{2n^2}\sum_{i=1}^n \sum_{j=1}^n \left[\E_{\lambda'_1} m_{ji}(\lambda'_1) + \E_{\lambda_1} m_{ij}(\lambda_1) \right]\\
        &= \frac{1}{n^2}\sum_{i=1}^n \sum_{j=1}^n \E_{\lambda_1} m_{ij}(\lambda_1)\\
        &= \frac{1}{n}\sum_{i=1}^n \left[ \frac{1}{n} \sum_{j=1}^n \E_{\lambda_1} m_{ij}(\lambda_1) \right] \\
        &= \frac{1}{n}\sum_{i=1}^n \ell_i\,,
    \end{split}
\end{equation}
where
$$
\ell_i = \E_{\theta,j} \ell\left(\theta y_i + (1-\theta)y_j , f(\theta x_i + (1-\theta) x_j)\right)\,,\quad \theta \sim \text{Beta}_{[\frac{1}{2},1]}(\alpha,\alpha)\,,\quad j \sim \text{Unif}([n])\,.
$$
We now easily see that $\tilde{x}_i$ and $\tilde{y}_i$ defined in (\ref{eq:modif}) satisfy
$$
\begin{cases}
\tilde{x_i} = \E_{\theta,j} \left[\theta x_i + (1-\theta) x_j \right]\,,\\
\tilde{y_i} = \E_{\theta,j} \left[\theta y_i + (1-\theta) y_j \right]\,,
\end{cases}
$$
and furthermore that $\delta_i$ and $\epsilon_i$ defined in (\ref{eq:perturbations}) satisfy
$$
\begin{cases}
\tilde{\delta_i} = \theta x_i + (1-\theta) x_j - \E_{\theta,j} \left[\theta x_i + (1-\theta) x_j \right]\,,\\
\tilde{\epsilon_i} = \theta y_i + (1-\theta) y_j - \E_{\theta,j} \left[\theta y_i + (1-\theta) y_j \right]\,,
\end{cases}
$$
from which we deduce that $\E_{\theta,j} \delta_i = \E_{\theta,j} \epsilon_i=0$ and
$$
\ell_i = \E_{\theta,j} \ell\left(\tilde{y}_i + \tilde{\epsilon_i} , f(\tilde{x}_i + \tilde{\delta_i}) \right)\,.
$$
Plugging this equality back into (\ref{eq:technical}) concludes the proof.
\end{proof}

\subsection{Proof of Lemma~\ref{lem:corr}}

\begin{proof}
From the definition of $\delta_i$ in (\ref{eq:perturbations}), we easily get:
\begin{equation}\label{eq:technical2}
    \begin{split}
        \E_{\theta,j} \delta_i \delta_i^\top
        &= \E_{\theta} (\theta - \bar{\theta})^2 x_i x_i^\top + \E_{\theta}[(1-\theta)^2] \E_{j}[ x_j x_j^\top] + (1-\bar{\theta})^2\bar{x}\bar{x}^\top \\
        & \quad + \E_{\theta}[(\theta-\bar{\theta})(1-\theta)] \E_{j}[x_i x_j^\top + x_j x_i^\top] - (1-\bar{\theta})\E_{\theta}[1-\theta] \E_{j}[x_j \bar{x}^\top + \bar{x} x_j^\top] \\
        &= \sigma^2 x_i x_i^\top + \gamma^2 \E_{j}[x_j x_j^\top] + (1-\bar{\theta})^2 \bar{x}\bar{x}^\top - \sigma^2 (x_i \bar{x}^\top + \bar{x} x_i^\top) - 2 (1-\bar{\theta})^2 \bar{x}\bar{x}^\top \\
        &= \sigma^2( x_i x_i^\top - x_i \bar{x}^\top - \bar{x} x_i^\top) + \gamma^2(\Sigma_{xx} + \bar{x}\bar{x}^\top) - (1-\bar{\theta})^2 \bar{x} \bar{x}^\top\\
        &= \sigma^2( x_i- \bar{x})( x_i- \bar{x})^\top  + \gamma^2 \Sigma_{xx}\,,
    \end{split}
\end{equation}
where we used the independence between $\theta$ and $j$ in the first equality; for the second, the facts that $\E_{\theta} (\theta - \bar{\theta})^2=\sigma^2$, $\E_{\theta}[(1-\theta)^2] = \sigma^2 + (1-\bar{\theta})^2= \gamma^2$, $\E_{\theta}[(\theta-\bar{\theta})(1-\theta)] = \bar{\theta}^2 - \E_{\theta}[\theta^2] = -\sigma^2$, and $\E_{\theta}[1-\theta]=1-\bar{\theta}$; for the third, we reorganized the terms and used the equality $\E_j x_j x_j^\top = \Sigma_{xx} + \bar{x}\bar{x}^\top$ by definition of the empirical covariance matrix $\Sigma_{xx}$; the last equality is obtained by reorganizing the terms and using the definition of $\gamma^2$. In order to write this covariance matrix in terms of modified inputs, we notice that by definition (\ref{eq:modif}) we have $x_i - \bar{x} = (\tilde{x}_i - \bar{x})/\bar{\theta}$ and $\E_j \tilde{x}_j = \E_j x_j = \bar{x}$, which implies that the empirical covariance matrix of the modified inputs is $\Sigma_{\tilde{x}\tilde{x}} = \bar{\theta}^2 \Sigma_{xx}$. Combining these equalities with (\ref{eq:technical2}) gives the first equality in (\ref{eq:cov}). The two other equalities can be proved exactly the same way.
\end{proof}

\subsection{Proof of \Cref{thm:mixupreg}}

\begin{proof}
Given a modified input/output pair $(\tilde{x},\tilde{y})\in\X\times\Y$ and a function $f\in\hh$, the second-order Taylor approximation of the loss $G(\tilde{x},\tilde{y}) = \ell(\tilde{y},f(\tilde{x}))$ is, for any $(\delta,\epsilon)\in\X\times\Y$:
\begin{equation}
    \begin{split}
G_Q(\tilde{x}+\delta,\tilde{y}+\epsilon)= G(\tilde{x},\tilde{y}) &+ \nabla_{\tilde{x}} G(\tilde{x},\tilde{y}) \delta + \nabla_{\tilde{y}} G(\tilde{x},\tilde{y}) \epsilon\\
&+ \frac{1}{2} \delta^\top  \nabla^2_{\tilde{x}\tilde{x}} G(\tilde{x},\tilde{y}) \delta
+ \frac{1}{2} \epsilon^\top  \nabla^2_{\tilde{y}\tilde{y}} G(\tilde{x},\tilde{y}) \epsilon + \epsilon^\top  \nabla^2_{\tilde{y}\tilde{x}} G(\tilde{x},\tilde{y}) \delta\,.
\end{split}
\end{equation}
Using this quadratic approximation at each training point $i\in[n]$ in (\ref{eq:mixupERM}), and using the fact that $\E_{\theta,j} \delta_i = \E_{\theta,j} \epsilon_i = 0$, we get
\begin{equation}
    \begin{split}
\E_{\theta,j} G_Q(\tilde{x}_i+\delta_i,\tilde{y}_i+\epsilon_i)= &G(\tilde{x}_i,\tilde{y}_i)
+ \frac{1}{2} \scal{\E_{\theta,j}\delta_i\delta_i^\top}{\nabla^2_{\tilde{x}\tilde{x}} G(\tilde{x}_i,\tilde{y}_i)}\\
&+ \frac{1}{2} \scal{\E_{\theta,j}\epsilon_i \epsilon_i^\top}{\nabla^2_{\tilde{y}\tilde{y}} G(\tilde{x}_i,\tilde{y}_i)}
+ \scal{\E_{\theta,j}\epsilon_i \delta_i^\top}{\nabla^2_{\tilde{y}\tilde{x}} G(\tilde{x}_i,\tilde{y}_i)}\,,
\end{split}
\end{equation}
which we can rewrite as follows by expressing the derivatives of $G(x,y) = \ell(y,f(x))$ in terms of derivatives of $\ell(y,u)$ and $f(x)$:
\begin{equation}
    \begin{split}
\E_{\theta,j} \ell_Q(\tilde{y}_i+\epsilon_i, f&(\tilde{x}_i+\delta_i))= \ell( \tilde{y}_i , f(\tilde{x}_i))\\
&+ \frac{1}{2} \scal{\E_{\theta,j}\delta_i\delta_i^\top}{\nabla f(\tilde{x}_i)^\top \nabla^2_{uu}\ell(\tilde{y}_i,f(\tilde{x}_i)) \nabla f(\tilde{x}_i) + \nabla_u\ell(\tilde{y}_i,f(\tilde{x}_i))\nabla^2 f(\tilde{x}_i)}\\
&+ \frac{1}{2} \scal{\E_{\theta,j}\epsilon_i \epsilon_i^\top}{\nabla^2_{yy}\ell(\tilde{y}_i,f(\tilde{x}_i))}
+ \scal{\E_{\theta,j}\epsilon_i \delta_i^\top}{\nabla^2_{yu}\ell(\tilde{y}_i,f(\tilde{x}_i))\nabla f(\tilde{x}_i)}\,.
\end{split}
\end{equation}
Replacing the expectations in this equation by their values given by Lemma~\ref{lem:corr} gives:
\begin{equation}\label{eq:technical3}
    \begin{split}
\E_{\theta,j} \ell_Q(\tilde{y}_i+\epsilon_i, f(\tilde{x}_i+\delta_i))
&= \ell( \tilde{y}_i , f(\tilde{x}_i)) \\
&+ \frac{1}{2} \scal{\Sigma_{\tilde{x}\tilde{x}}^{(i)}}{\nabla f(\tilde{x}_i)^\top \nabla^2_{uu}\ell(\tilde{y}_i,f(\tilde{x}_i)) \nabla f(\tilde{x}_i)}\\
&+ \frac{1}{2} \scal{\Sigma_{\tilde{x}\tilde{x}}^{(i)}}{ \nabla_u\ell(\tilde{y}_i,f(\tilde{x}_i))\nabla^2 f(\tilde{x}_i)}\\
&+ \frac{1}{2} \scal{\Sigma_{\tilde{y}\tilde{y}}^{(i)}}{\nabla^2_{yy}\ell(\tilde{y}_i,f(\tilde{x}_i))}\\
&+ \scal{\Sigma_{\tilde{y}\tilde{x}}^{(i)}}{\nabla^2_{yu}\ell(\tilde{y}_i,f(\tilde{x}_i))\nabla f(\tilde{x}_i)}\,.
\end{split}
\end{equation}
We now use the following fact, true for any square symmetric and invertible matrices $A$ and $C$ and rectangular matrices $B$ and $Y$ (such that the matrix multiplications below make sense):
\begin{equation}\label{eq:matrixtrick}
\scal{A}{Y^\top C Y} - 2\scal{B}{Y} = \scal{A}{(Y-Z)^\top C (Y-Z)} - \scal{A^{-1}}{B^\top C^{-1} B}\,,
\end{equation}
where $Z = C^{-1} B A^{-1}$, to combine the second and fifth terms together. Indeed, the fifth term (\ref{eq:technical3}) can be rewritten as
$$
\scal{\Sigma_{\tilde{y}\tilde{x}}^{(i)}}{\nabla^2_{yu}\ell(\tilde{y}_i,f(\tilde{x}_i))\nabla f(\tilde{x}_i)} = \scal{\nabla^2_{uy}\ell(\tilde{y}_i,f(\tilde{x}_i))\Sigma_{\tilde{y}\tilde{x}}^{(i)}}{\nabla f(\tilde{x}_i)\,,} \,,
$$
so plugging into (\ref{eq:matrixtrick}) the following matrices:
$$
\begin{cases}
A &= \Sigma_{\tilde{x}\tilde{x}}^{(i)}\,,\\
B &= - \nabla^2_{uy}\ell(\tilde{y}_i,f(\tilde{x}_i)) \Sigma_{\tilde{y}\tilde{x}}^{(i)}\,,\\
C &= \nabla^2_{uu}\ell(\tilde{y}_i,f(\tilde{x}_i))\,,\\
Y &= \nabla f(\tilde{x}_i)\,,
\end{cases}
$$
gives
\begin{equation}\label{eq:technical4}
    \begin{split}
        \frac{1}{2} &\scal{\Sigma_{\tilde{x}\tilde{x}}^{(i)}}{\nabla f(\tilde{x}_i)^\top \nabla^2_{uu}\ell(\tilde{y}_i,f(\tilde{x}_i)) \nabla f(\tilde{x}_i)} + \scal{\Sigma_{\tilde{y}\tilde{x}}^{(i)}}{\nabla^2_{yu}\ell(\tilde{y}_i,f(\tilde{x}_i))\nabla f(\tilde{x}_i)} \\
        = &\frac{1}{2} \scal{\Sigma_{\tilde{x}\tilde{x}}^{(i)}}{\left(\nabla f(\tilde{x}_i) - J^{(i)}\right)^\top \nabla^2_{uu}\ell(\tilde{y}_i,f(\tilde{x}_i)) \left(\nabla f(\tilde{x}_i) - J^{(i)}\right)} \\
        &-\frac{1}{2} \scal{\Sigma_{\tilde{y}\tilde{x}}^{(i)}\left(\Sigma_{\tilde{x}\tilde{x}}^{(i)}\right)^{-1} \Sigma_{\tilde{x}\tilde{y}}^{(i)}}{\nabla^2_{yu}\ell(\tilde{y}_i,f(\tilde{x}_i))
        \nabla^2_{uu}\ell(\tilde{y}_i,f(\tilde{x}_i))^{-1}
        \nabla^2_{uy}\ell(\tilde{y}_i,f(\tilde{x}_i)) }\, \\
\end{split}
\end{equation}
where $J^{(i)}$ is defined in (\ref{eq:Ji}).
Theorem~\ref{thm:mixupreg} then follows by merging the second and fifth terms in (\ref{eq:technical3}) using (\ref{eq:technical4}), and summing over $i$.
\end{proof}

\subsection{Proof of Corollary~\ref{cor:CEloss}}

\begin{proof}
For the definition (\ref{eq:loss}) of the cross-entropy loss $\ell^{\text{CE}}(y,u)$, we easily get:
\begin{equation*}
    \begin{split}
    \nabla_y \ell^{\text{CE}}(y,u) &= -u^\top\,,\\
    \nabla_u \ell^{\text{CE}}(y,u) &= (\Scal(u) - y)^\top \,, \\
    \nabla^2_{yy} \ell^{\text{CE}}(y,u) &= \bzero_c \,,\\
    \nabla^2_{yu} \ell^{\text{CE}}(y,u) &= - \bI_c \,,\\
    \nabla^2_{uu} \ell^{\text{CE}}(y,u) &= H(u)\,.
    \end{split}
\end{equation*}
Plugging these results back in the four regularization terms in \Cref{thm:mixupreg} we conclude the proof.
\end{proof}

\subsection{Proof of Corollary~\ref{cor:logisticloss}}

\begin{proof}
For the definition (\ref{eq:lrloss}) of the logistic regression loss $\ell^{\text{LR}}(y,u)$, we easily get:
\begin{equation*}
    \begin{split}
    \nabla_y \ell^{\text{LR}}(y,u) &= -u\,,\\
    \nabla_u \ell^{\text{LR}}(y,u) &= s(u) - y \,, \\
    \nabla^2_{yy} \ell^{\text{LR}}(y,u) &= 0 \,,\\
    \nabla^2_{yu} \ell^{\text{LR}}(y,u) &= - 1 \,,\\
    \nabla^2_{uu} \ell^{\text{LR}}(y,u) &= v(u)\,.
    \end{split}
\end{equation*}
Plugging these results back in the four regularization terms in \Cref{thm:mixupreg} we conclude the proof.
\end{proof}

\subsection{Proof of Corollary~\ref{cor:SEloss}}

\begin{proof}
For the definition (\ref{eq:loss}) of the squared error loss $\ell^{\text{SE}}(y,u)$, we easily get:
\begin{equation*}
    \begin{split}
    \nabla_y \ell^{\text{SE}}(y,u) &= (y-u)^\top\,,\\
    \nabla_u \ell^{\text{SE}}(y,u) &= (u - y)^\top \,, \\
    \nabla^2_{yy} \ell^{\text{SE}}(y,u) &= \nabla^2_{uu} \ell^{\text{SE}}(y,u) = \bI_c \,,\\
    \nabla^2_{yu} \ell^{\text{SE}}(y,u) &= - \bI_c \,.
    \end{split}
\end{equation*}
Plugging these results back in the 4 regularization terms in \Cref{thm:mixupreg} proves (\ref{eq:QmixupSE}).

When $f$ is a linear function with intercept of the form $f_{W,b}(x) = Wx+b$, then we first note that $\ell^{SE}(y, f_{W,b}(x))$ is a quadratic function of $(x,y)$, so the second-order Taylor approximation (\ref{eq:taylor}) is exact in that case:
$\ell^{\text{SE}(i)}_Q(y, f_{W,b}(x)) = \ell^{\text{SE}}(y, f_{W,b}(x))$ for any $i\in[n]$ and $(x,y)\in\X\times\Y$, and consequently:
$$
\forall (W,b)\in\R^{c\times d}\times\R^c\,,\quad\Ex^{\text{Mixup}}_Q(f_{W,b}) = \Ex^{\text{Mixup}}(f_{W,b})\,.
$$

Applying (\ref{eq:QmixupSE}) to the case of a linear function $f_{W,b}$ gives us immediately $R_2^{\text{SE}}(f_{W,b}) = 0$, because $\nabla^2f_{W,b} = 0$. For the first regularization term, we compute
\begin{equation}\label{eq:technical10}
    \begin{split}
        R_1^{\text{SE}}(f_{W,b}) &= \frac{1}{2n} \sum_{i=1}^n \| \nabla f(\tilde{x}_i) - J^{(i)} \|^2_{\Sigma_{\tilde{x}\tilde{x}}^{(i)}}\\
        &= \frac{1}{2n} \sum_{i=1}^n \| W - J^{(i)} \|^2_{\Sigma_{\tilde{x}\tilde{x}}^{(i)}} \\
        &= \frac{1}{2n} \sum_{i=1}^n \left(\scal{W^\top W}{\Sigma_{\tilde{x}\tilde{x}}^{(i)}} -2  \scal{W}{\Sigma_{\tilde{y}\tilde{x}}^{(i)}}\right)  + C \\
        &= \frac{1}{2n} \sum_{i=1}^n \left[ \frac{\gamma^2}{n\bar{\theta}^2} \sum_{j=1}^n \|W (\tilde{x}_j-\bar{x}) - (\tilde{y}_j - \bar{y})\|^2 + \frac{\sigma^2}{\bar{\theta}^2}\|W(\tilde{x}_i-\bar{x}) - (\tilde{y}_i-\bar{y})\|^2\right] +C\\
        &= \frac{\sigma^2 + \gamma^2}{2n\bar{\theta}^2} \sum_{i=1}^n \|W(\tilde{x}_i -\bar{x})- (\tilde{y}_i-\bar{y})\|^2 +C\\
        &= \frac{2\sigma^2 + (1-\bar{\theta})^2}{2n} \sum_{i=1}^n \|W(x_i -\bar{x})- (y_i-\bar{y})\|^2 +C\,.
    \end{split}
\end{equation}
As for the empirical risk term, we can also rewrite it as
\begin{equation}\label{eq:technical11}
    \begin{split}
      \frac{1}{n} \sum_{i=1}^n  \ell^{\text{SE}}( \tilde{y}_i , f_{W,b}(\tilde{x}_i))
      &= \frac{1}{n} \sum_{i=1}^n \| W \tilde{x}_i + b - \tilde{y}_i \|^2 \\
      &= \frac{1}{n} \sum_{i=1}^n \|W(\tilde{x}_i -\bar{x})- (\tilde{y}_i-\bar{y}) + (b - \bar{b})\|^2\\
      &= \frac{1}{n} \sum_{i=1}^n \|W(\tilde{x}_i -\bar{x})- (\tilde{y}_i-\bar{y})\|^2 + \|b - \bar{b}\|^2 \\
      &= \frac{\bar{\theta}^2}{n} \sum_{i=1}^n \|W({x}_i -\bar{x})- ({y}_i-\bar{y})\|^2 + \|b - \bar{b}\|^2 \\
    \end{split}
\end{equation}
Plugging (\ref{eq:technical10}) and (\ref{eq:technical11}) into (\ref{eq:QmixupSE}) finally gives (\ref{eq:mixupOLS}).

To see that the minimizer of (\ref{eq:mixupOLS}) is the standard MOLS solution, we notice that the obvious solution for $b$ is $b=\bar{b}$, which is the intercept of MOLS, while the solution for $W$ should minimize the sum of squared errors over centered points, which is exactly what MOLS does.
\end{proof}

\subsection{Proof of Proposition~\ref{prop:label_smoothing_logistic_reg}}

\begin{proof}
We first start by deriving a dual formulation for~(\ref{eq:primal_logistic_reg_with_ls}). The derivation for~(\ref{eq:primal_logistic_reg_without_ls}) follows along the same lines, up to the replacement of $\tilde{y}_i$ by $y_i$.

Introducing primal variables $u_i \in \R^c$ for $i \in [n]$ with equality constraints $W x_i = u_i$ and the dual variables $\xi_i \in \R^c$, we obtain the following Lagrangian (see~\citet{boyd2004convex}):
\begin{eqnarray*}
\Lcal(\{u_i, \xi_i\}_{i=1}^n, W) &=&
\frac{1}{n} \sum_{i=1}^n \big\{  \ell^{\text{CE}}(\tilde{y}_i, u_i)
+ \xi_i^\top ( W x_i - u_i) \big\}\\
&=&
\frac{1}{n} \sum_{i=1}^n \Big\{ 
\log\Big(
\sum_{j=1}^c e^{(u_i)_j}
\Big) - (\xi_i + \tilde{y}_i)^\top u_i
\Big\}
+
\scal{W}{\frac{1}{n} \sum_{i=1}^n \xi_i x_i^\top} \,.
\end{eqnarray*}
We recall that the entropy is concave and is the Fenchel conjugate, up to a sign flip, of the log-sum-exp function
\begin{equation}\label{eq:fenchel_log_sum_exp}
    \Hcal(p) = -\max_{t \in \R^c} \Big\{p^\top t - \log\Big(\sum_{j=1}^c e^{t_j}
    \Big)
    \Big\},
\end{equation}
as for instance detailed in Example 5.5 of~\citep{boyd2004convex}. We therefore derive the dual function using~(\ref{eq:fenchel_log_sum_exp})
\begin{equation*}
\min_{\{u_i\}_{i=1}^n, W} \Lcal(\{u_i, \xi_i\}_{i=1}^n, W)
=
\begin{cases}
\frac{1}{n}\sum_{i=1}^n \Hcal(\xi_i + \tilde{y}_i)\ \ \text{if}\ \ \sum_{i=1}^n \xi_i x_i^\top = 0
\ \text{and}\ \xi_i + \tilde{y}_i \in \Delta_c,
\\
-\infty\ \ \text{otherwise}
\end{cases}
\end{equation*}
and the dual problem
\begin{equation}\label{eq:dual_logitstic_reg}
\max_{\{\nu_i\}_{i=1}^n}
\frac{1}{n}\sum_{i=1}^n \Hcal(\nu_i)
\ \ \text{subject to}\ \
 \nu_i \in \Delta_c,\ i\in[n],
\ \ \text{and}\ \ \sum_{i=1}^n (\nu_i - \tilde{y}_i) x_i^\top = 0
\end{equation}
where we have made the change of variables $\nu_i = \xi_i + \tilde{y}_i$. The dual problem for (\ref{eq:primal_logistic_reg_without_ls}) is identical to~(\ref{eq:dual_logitstic_reg}) up to the replacement of $\tilde{y}_i$ by $y_i$.

Recalling the definitons in~(\ref{eq:prediction_softmax_without_with_ls}) and exploiting the first-order optimality conditions of~(\ref{eq:primal_logistic_reg_without_ls}) and~(\ref{eq:primal_logistic_reg_with_ls}), we have
\begin{equation*}
    \sum_{i=1}^n (p_i - y_i) x_i^\top = 0
\ \ \text{and}\ \
    \sum_{i=1}^n (\tilde{p}_i - \tilde{y}_i) x_i^\top = 0,
\end{equation*}
so that $\{\tilde{p}_i\}_{i=1}^n$ is feasible for the dual problem~(\ref{eq:dual_logitstic_reg}).
Since strong duality applies~\citep{boyd2004convex}, it also holds that $\{\tilde{p}_i\}_{i=1}^n$ maximize~(\ref{eq:dual_logitstic_reg}).

Let us consider $\{q_i\}_{i=1}^n$ defined for $i \in [n]$ by
\begin{equation*}
q_i = \bar{\theta} p_i + (1 - \bar{\theta}) \bar{y}.
\end{equation*}
We can easily observe that $q_i \in \Delta_c$ as convex combination of $p_i, \bar{y} \in \Delta_c$ and that
\begin{equation*}
 \sum_{i=1}^n (q_i - \tilde{y}_i) x_i^\top
 =
\bar{\theta}\sum_{i=1}^n ({p}_i - {y}_i) x_i^\top 
+
(1-\bar{\theta}) \sum_{i=1}^n (\bar{y} - \bar{y}) x_i^\top
=0.
\end{equation*}
This implies that $\{q_i\}_{i=1}^n$ is feasible for (\ref{eq:dual_logitstic_reg}) and we have
\begin{equation*}
    \frac{1}{n}\sum_{i=1}^n \Hcal(q_i)
    \leq 
    \frac{1}{n}\sum_{i=1}^n \Hcal(\tilde{p}_i).
\end{equation*}
We get the advertised result by using the concavity of $\Hcal$ so that $\bar{\theta}\Hcal(p_i) + (1-\bar{\theta}) \Hcal(\bar{y}) \leq \Hcal(q_i)$.
\end{proof}

\section{Experiments}\label{sec:app_exp}
\subsection{CIFAR-10 and CIFAR-100}
To learn on CIFAR-10, CIFAR-100 we use ResNet-34 and LeNet.
For both architectures the optimizer used for training is SGD with momentum $0.9$ for $200$ epochs 
with mini-batch size $128$, weight-decay $5 \times 10^{-4}$.
For ResNet-34 the learning rate is $0.1$ reduced by a factor $10$ at epoch $60, 120, 160$.
For LeNet the learning rate is $0.01$ reduced by a factor $10$ at epoch $100$.

\subsection{ImageNet}
To learn on the ImageNet we use ResNet-50.
The optimizer used for training is SGD with Nesterov and momentum $0.9$ for $200$ epochs 
with mini-batch size $4096$ ($32$ numbers of cores $\times 128$ per core batch-size), weight-decay $5 \times 10^{-5}$.
The learning rate is $1.6$ reduced by a factor $10$ at epoch $66, 133, 177$.

\subsection{Two Moons with Random Features}
\begin{figure*}[ht]
\hfill
    \minipage{0.33\textwidth}
    \includegraphics[width=\linewidth]{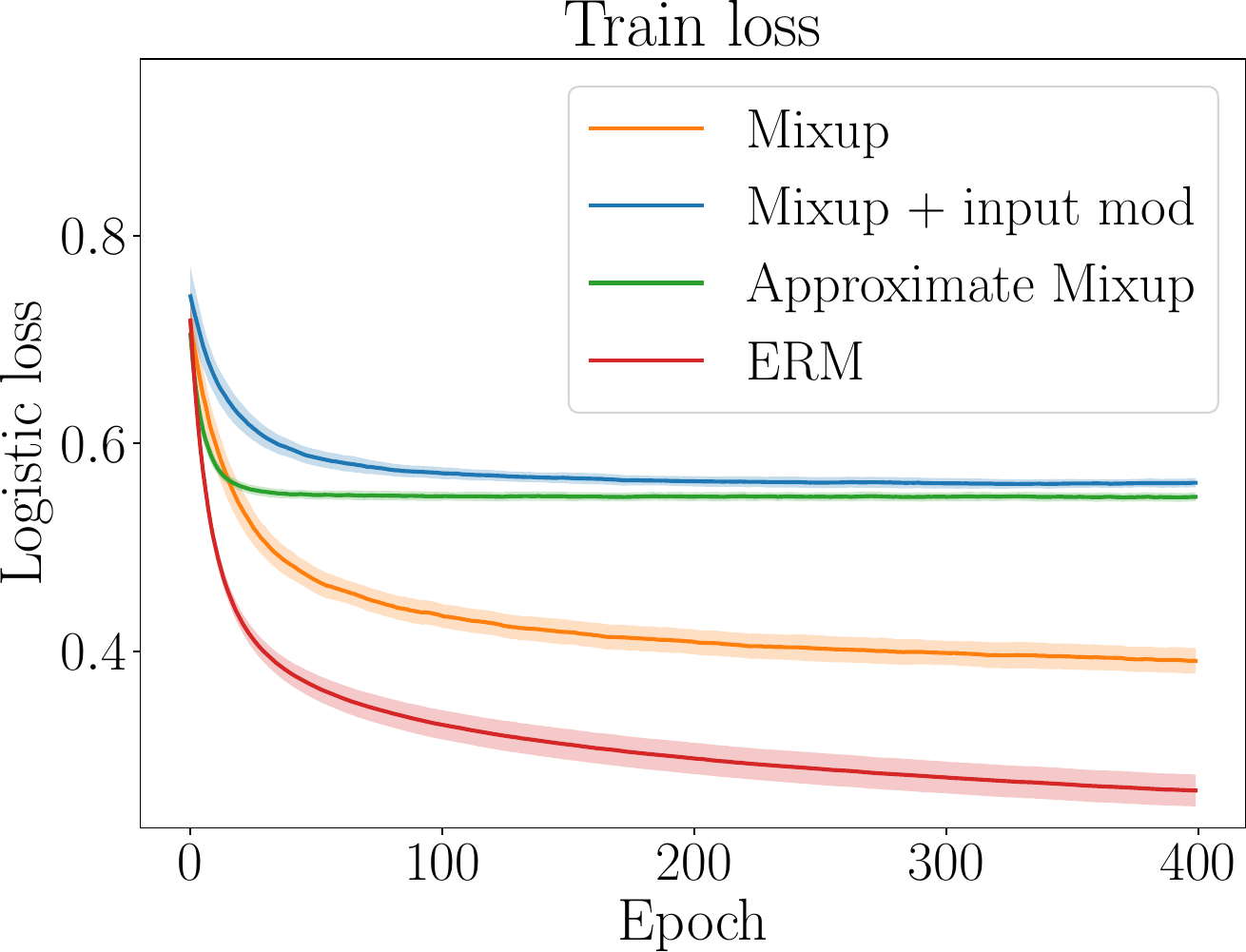}
    \endminipage\hfill
    \minipage{0.33\textwidth}
    \includegraphics[width=\linewidth]{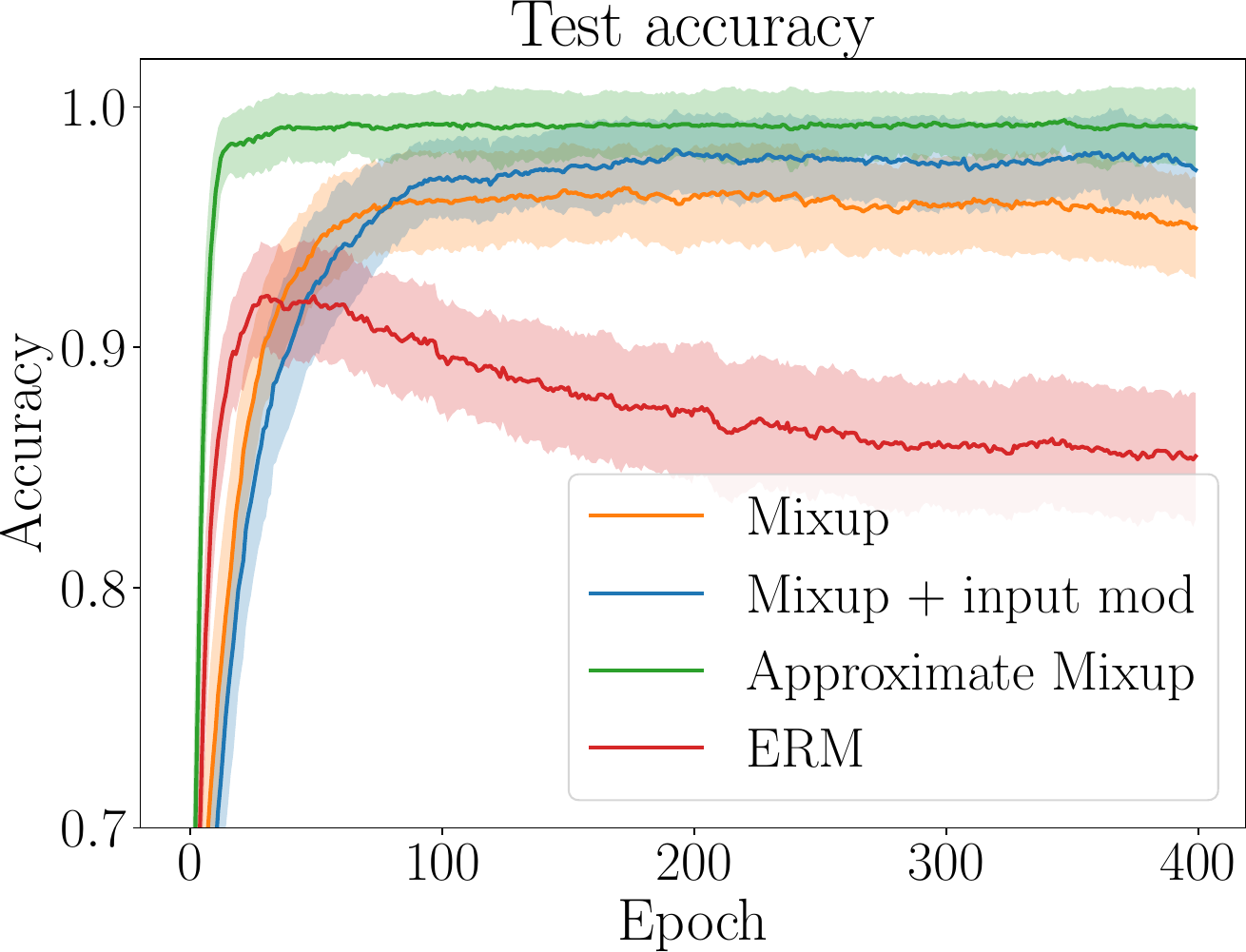}
    \endminipage\hfill 
    \minipage{0.33\textwidth}
    \includegraphics[width=\linewidth]{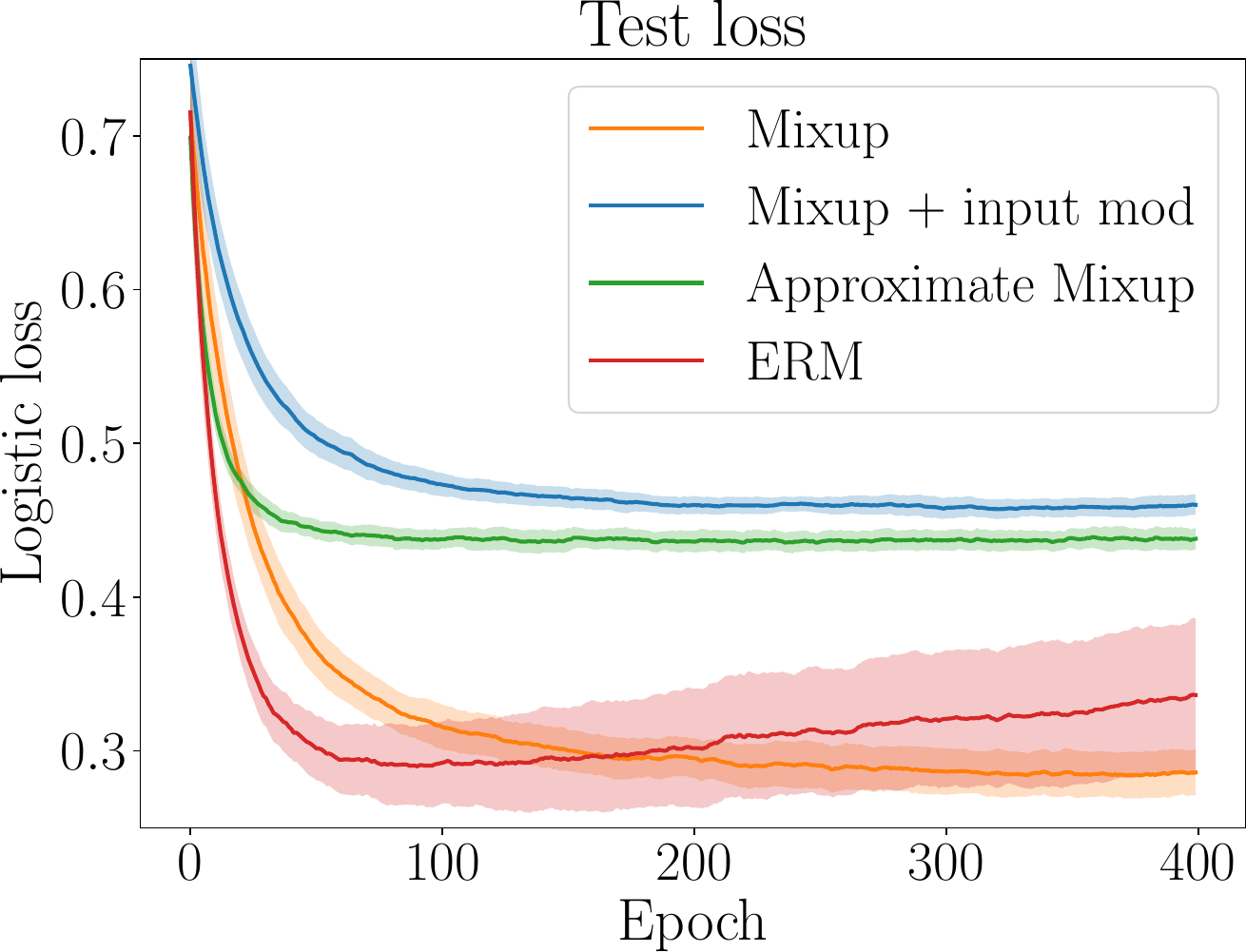}
    \endminipage\hfill
\caption{
        \small{From left to right: train loss, 
        train and test accuracy during optimization of a 
        logistic regression model trained on the noisy two-moon problem with Mixup, Mixup (rescaled), 
        approximate Mixup and ERM risks.}
        }\label{fig:mess_vs_approx_acc_twomoons}
\end{figure*}

\begin{figure*}[h]
\centering
    \minipage{0.42\textwidth}
    \includegraphics[width=\linewidth]{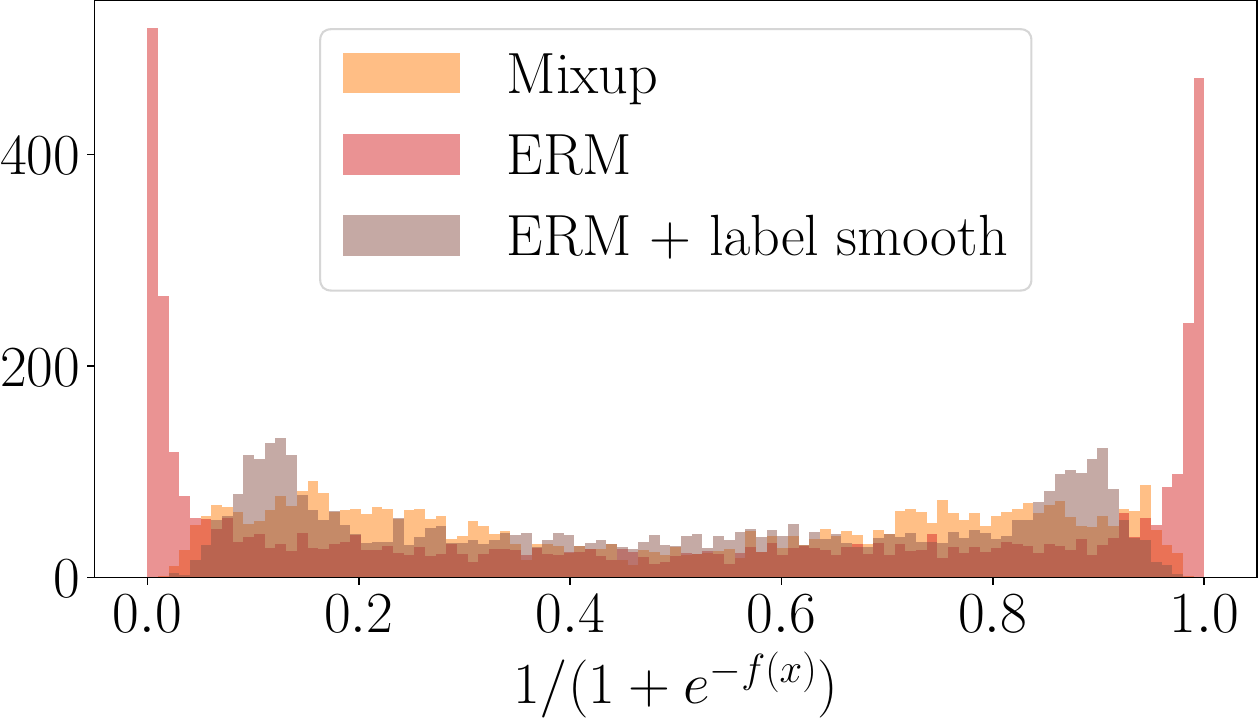}
    \endminipage
    \hspace{1cm}
\minipage{0.42\textwidth}
    \includegraphics[width=\linewidth]{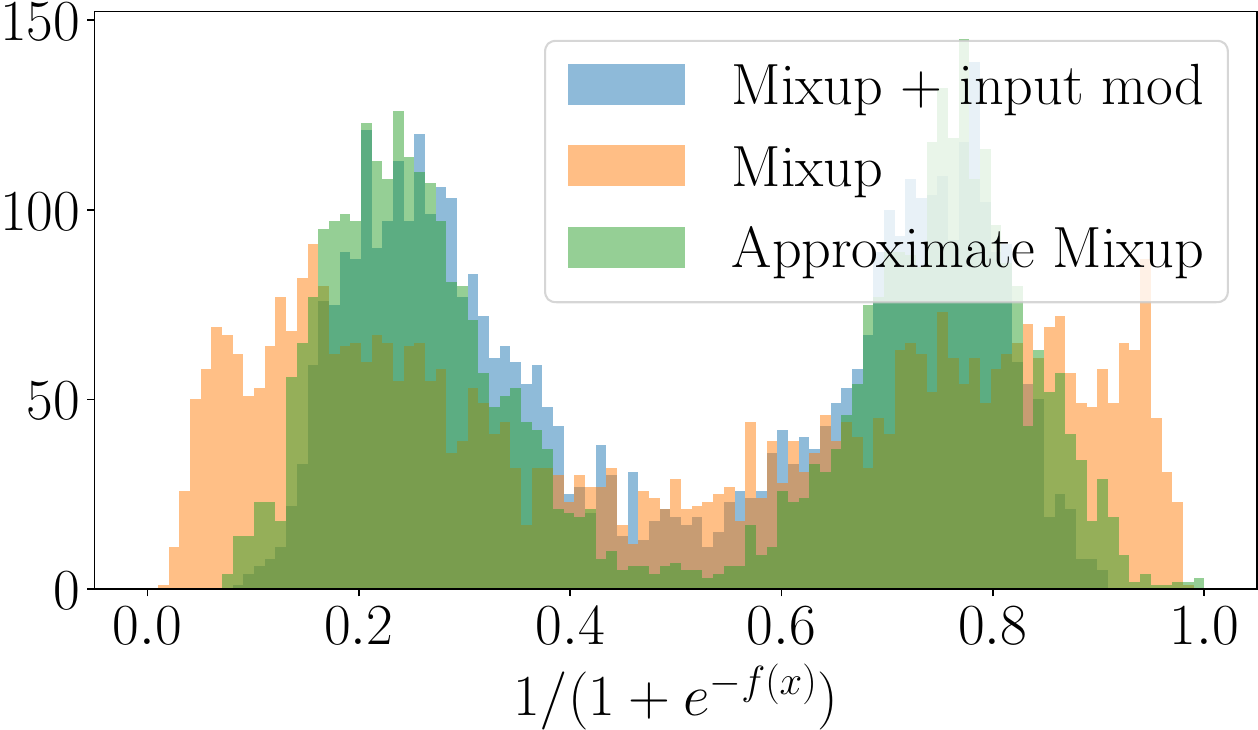}
    \endminipage\hfill
    \caption{\small{Histograms of confidence of predictions on test points, for models trained with different techniques.}}\label{fig:hist_twomoon}
\end{figure*}

We report in \Cref{fig:mess_vs_approx_acc_twomoons} and \Cref{fig:hist_twomoon} some more results on a synthetic binary classification problem (noisy two-moon problem),
where we train a logistic regression model with random Fourier features~\citep{rahimi2008random}.
This allowed us to get rid of convergence issues due to the convexity of the problem, but still work with nonlinear models of the input points. 
For each experimental result we report mean and 95\% confidence interval using 30 repetitions over 30 different instances of the data.

What we notice from these results is similar behaviors as reported for CIFAR-10, CIFAR-100, ImageNet in the main paper. 

\paragraph{Data generation.} 
To generate the data we use the \texttt{sklearn.datasets.make\_moons} function from the \texttt{scikit-learn} library.
We create $n=300$ points with \emph{noise}$= 0.01$, and split them in $50\%$ for train and $50\%$ for test. We then randomly flip $20\%$ of the training labels to make the learning task more difficult. We repeat this pipeline 30 times for 30 different random seeds.

\paragraph{Function space.}
Let $M=1000$, $w \in \R^M$ and $\phi : \R^d \rightarrow \R^M$ be the feature map defined as $\phi(x) = \frac{1}{\sqrt{M}}\cos(Sx + B)$, where $\cos : \R^M \rightarrow \R^M$ is the element-wise \emph{cosine} function, $S\in\R^{M \times d}$ is the random matrix s.t. $S_{i,j} \sim \mathcal{N}(0, \sigma^2), \forall i \in [M], j\in [d]$ with $\sigma = 10$, and $B \in \R^M$ s.t. $B_i \sim \text{Unif}(0, 2\pi), \forall i \in [M]$.
The space of candidate solutions $\hh$ we consider is the class of functions of the form $f(x) = w^\top \phi(x)$.

\paragraph{Optimization.}
To minimize any functional we use stochastic gradient descent with mini-batching, with mini-batch size $b = 50$ and step-size $\gamma = 5$.

\paragraph{Mixup hyperparameter.} We consider the Beta distribution in Mixup and its approximation to be Beta$(\alpha, \alpha)$ with $\alpha = 1$.

\end{document}